\RequirePackage[svgnames]{xcolor}

\documentclass[11pt,letterpaper]{mystyle}
\usepackage[all]{hypcap}
\usepackage[svgnames]{xcolor}
\usepackage[comma,authoryear,compress]{natbib}
\usepackage{hyperref}[citecolor=lightblue]
\usepackage{amsmath}
\usepackage{etoolbox}

\hypersetup{
    colorlinks = true,
    citecolor = {YaleBlue},
}

\usepackage{algorithm}
\usepackage{algorithmicx}
\usepackage{algpseudocode}
\usepackage{microtype}
\usepackage{graphicx}
\expandafter\def\csname ver@subfig.sty\endcsname{}
\usepackage{booktabs} %
\usepackage{float}
\usepackage{bigstrut}

\usepackage{comment}
\usepackage{amsmath}
\usepackage{amssymb}
\usepackage{mathtools}
\usepackage{amsthm}
\usepackage{mathrsfs}
\usepackage{nicefrac}
\usepackage{dsfont}
\usepackage{enumitem}
\usepackage{subcaption}
\usepackage{graphicx,subfig}
\usepackage{cleveref}
\usepackage{bxcoloremoji}
\usepackage{float}
\usepackage{algpseudocode}

\usepackage[utf8]{inputenc} %
\usepackage[T1]{fontenc}    %
\usepackage{url}            %
\usepackage{booktabs}       %
\usepackage{amsfonts}       %
\usepackage{nicefrac}       %
\usepackage{microtype}      %
\usepackage{graphicx}
\usepackage{amssymb}
\usepackage{fdsymbol}
\usepackage{wrapfig}
\usepackage{lipsum}
\usepackage{enumitem}
\usepackage{stackengine}
\usepackage[font=small,labelfont=bf]{caption}
\usepackage{color}
\usepackage{adjustbox}

\usepackage{rotating}
\usepackage{makecell}

\usepackage{multirow}

\newcommand{\x}{\mathbf{x}}

\definecolor{blanchedalmond}{rgb}{1.0, 0.92, 0.8}
\definecolor{carmine}{rgb}{0.59, 0.0, 0.09}
\definecolor{lightblue}{rgb}{0.22,0.45,0.70}%

\newtheorem{theorem}{Theorem}[section]

\newtheorem{definition}[theorem]{Definition}

\renewcommand{\mathbf}{\boldsymbol}

\makeatletter
\def\Ddots{\mathinner{\mkern1mu\raise\p@
\vbox{\kern7\p@\hbox{.}}\mkern2mu
\raise4\p@\hbox{.}\mkern2mu\raise7\p@\hbox{.}\mkern1mu}}
\makeatother

\definecolor{amaranth}{rgb}{0.9, 0.17, 0.31}
\definecolor{antiquebrass}{rgb}{0.8, 0.58, 0.46}
\definecolor{antiquefuchsia}{rgb}{0.57, 0.36, 0.51}
\definecolor{chromeyellow}{rgb}{0.31, 0.47, 0.26}

\newtcolorbox{AIbox}[2][]{aibox,title=#2,#1}
\definecolor{lightblue}{rgb}{0.22,0.45,0.70}%
\definecolor{Gray}{gray}{0.95}
\definecolor{Cornsilk}{rgb}{1.0, 0.97, 0.86}

\usepackage{url}
\usepackage{xcolor}
\usepackage{pifont}
\usepackage{soul}

\usepackage{amsmath}

\usepackage[all]{hypcap}

\usepackage{algpseudocode}
\usepackage{authblk}   
\usepackage{amssymb}      
\usepackage{bbm}          
\usepackage{optidef}  
\usepackage[utf8]{inputenc}
\usepackage[T1]{fontenc}

\usepackage{mathpazo}
\usepackage[svgnames]{xcolor}
\usepackage{tcolorbox}

\usepackage{lipsum}

\newtcolorbox{simpleElegantQuote}{
    colback=AliceBlue!50!White,   
    colframe=RoyalBlue!75!Black,  
    boxrule=0.5pt,                
    arc=2mm,                     
    boxsep=4pt,                   
    left=10pt, right=10pt,        
    top=8pt, bottom=8pt,         
    fontupper=\itshape,          
}

\title{Memento: Fine-tuning LLM Agents without\\ Fine-tuning LLMs}

\runningtitle{Memento: Fine-tuning LLM Agents without Fine-tuning LLMs}

\author[1,2]{Huichi Zhou*}
\author[2]{Yihang Chen*}
\author[3]{Siyuan Guo}
\author[4]{Xue Yan}
\author[ ]{Kin Hei Lee}
\author[ ]{Zihan Wang}
\author[2]{Ka Yiu Lee}
\author[2]{Guchun Zhang}
\author[2]{Kun Shao}
\author[2]{Linyi Yang\dag}
\author[1]{Jun Wang\dag}

\affil[1]{AI Centre, UCL}
\affil[2]{Huawei Noah's Ark Lab, UK}
\affil[3]{Jilin University}
\affil[4]{Institute of Automation, CAS}

\correspondingauthor{Linyi Yang: yangly6@sustech.edu.cn, and Jun Wang: jun.wang@cs.ucl.ac.uk}

\begin{document}

\begin{abstract}
{\centering\section*{Abstract}}

In this paper, we introduce a novel learning paradigm for Adaptive Large Language Model (LLM) agents that eliminates the need for fine-tuning the underlying LLMs. Existing approaches are often either rigid, relying on static, handcrafted reflection workflows, or computationally intensive, requiring gradient updates of LLM model parameters. In contrast, our method enables low-cost continual adaptation via memory-based online reinforcement learning. We formalise this as a Memory-augmented Markov Decision Process (M-MDP), equipped with a neural case-selection policy to guide action decisions. Past experiences are stored in an episodic memory, either differentiable or non-parametric. The policy is continually updated based on environmental feedback through a memory rewriting mechanism, whereas policy improvement is achieved through efficient memory reading (retrieval). We instantiate our agent model in the deep research setting, namely \emph{Memento}, which attains top-1 on GAIA validation ($87.88\%$ Pass@$3$) and $79.40\%$ on the test set. It reaches $66.6\%$ F1 and $80.4\%$ PM on the DeepResearcher dataset, outperforming the state-of-the-art training-based method, while case-based memory adds $4.7\%$ to $9.6\%$ absolute points on out-of-distribution tasks. Our approach offers a scalable and efficient pathway for developing generalist LLM agents capable of continuous, real-time learning without gradient updates, advancing machine learning towards open-ended skill acquisition and deep research scenarios. The code is available at \url{https://github.com/Agent-on-the-Fly/Memento}.

\end{abstract}

\maketitle

\begin{figure}[h!]
  \centering
  \begin{subfigure}[t]{0.45\textwidth}
    \centering
    \includegraphics[width=0.9\linewidth,height=0.53\linewidth]{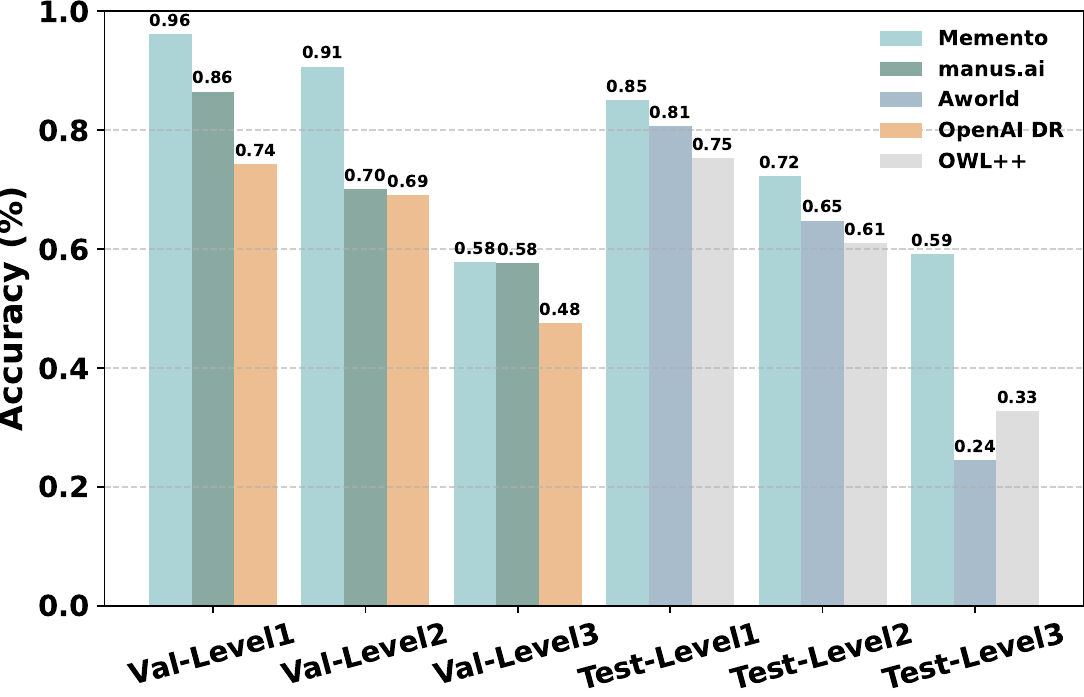}
    \caption{\emph{Memento} vs. Baselines on GAIA validation and test sets.}
    \label{fig:valtest}
  \end{subfigure}
  \hspace{1.2em}
  \begin{subfigure}[t]{0.45\textwidth}
    \centering    \includegraphics[width=0.9\linewidth,height=0.53\linewidth]{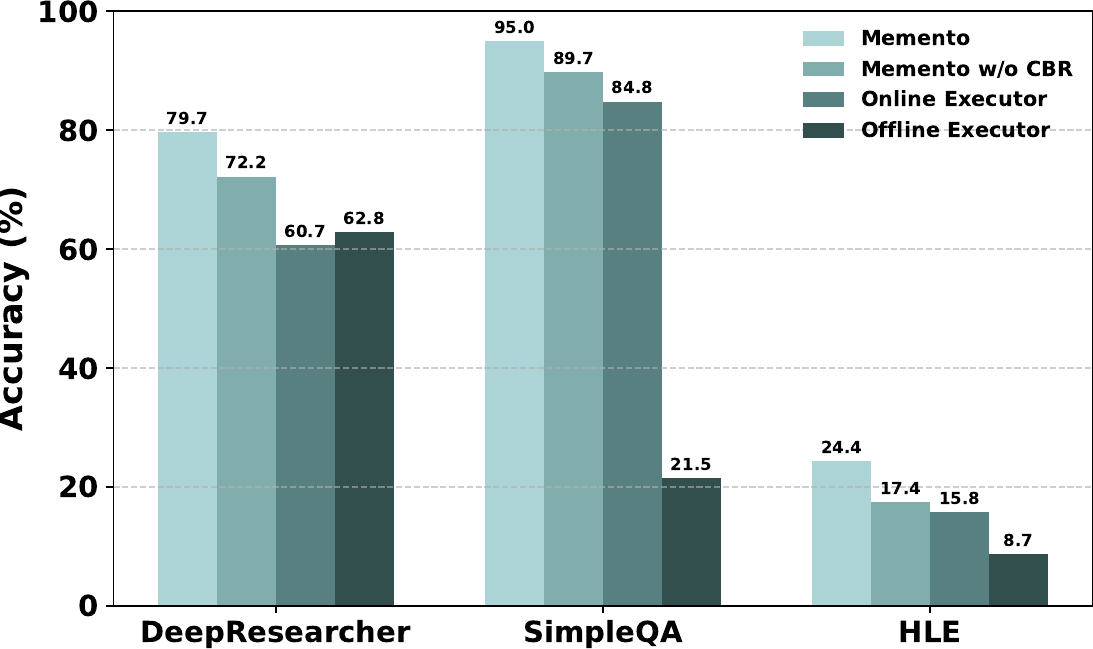}  
    \caption{Ablation study of \emph{Memento} across benchmarks.}
    \label{fig:tasks}
  \end{subfigure}
  \vspace{0.55em} 
  \begin{subfigure}[t]{0.45\textwidth}
    \centering    \includegraphics[width=0.9\linewidth,height=0.53\linewidth]{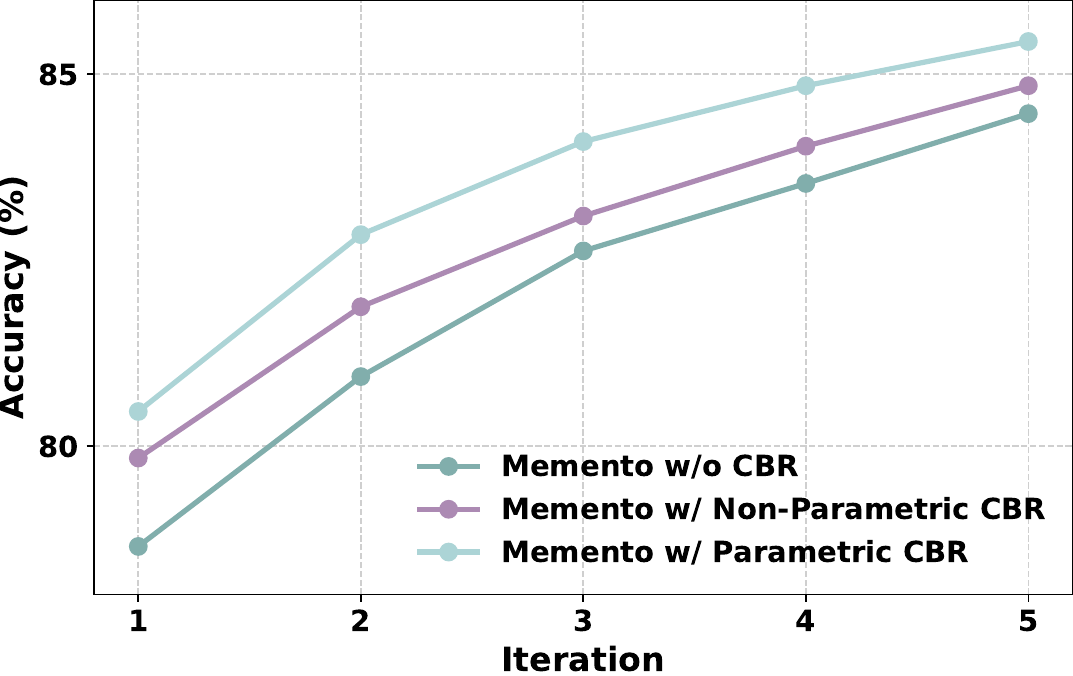}  
    \caption{Continual learning curves across memory designs.}
    \label{fig:iteration}
  \end{subfigure}
\hspace{1.2em}
  \begin{subfigure}[t]{0.45\textwidth}
    \centering
    \includegraphics[width=0.9\linewidth,height=0.53\linewidth]{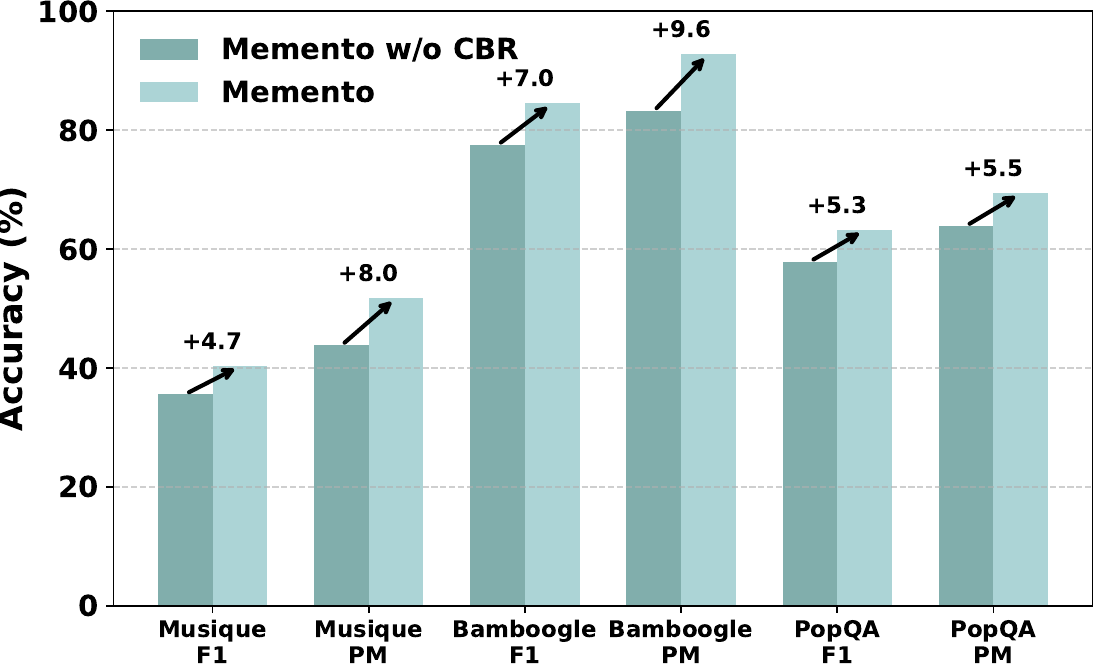}
    \caption{\emph{Memento}’s accuracy improvement on OOD datasets.}
    \label{fig:ood}
  \end{subfigure}

  \caption{Overview of \emph{Memento} evaluation across baselines, benchmarks, memory designs and generalisation.}
  \label{fig:combined}
\end{figure}

\section{Introduction}
\label{sec:intro}

A Large Language Model (LLM) agent refers to a system that leverages one or more LLMs to autonomously perform complex tasks through interaction, reasoning, and decision making, often with access to external tools, memory, or environments~\citep{christianos2023pangu,yang2025agentic}. Unlike passive LLMs that respond to prompts in isolation, LLM agents operate proactively and iteratively, guided by explicit goals. They are increasingly deployed as autonomous problem solvers~\citep{choudhary2021react, wei2022chain, yao2023} spanning various domains. Notable examples include deep research agents~\citep{openai2025deepresearch,google2025deepresearch,bytedance2025deerflow}, tool-enhanced execution systems~\citep{li2025search,zheng2025deepresearcher,qian2025toolrl}, and code generation agents~\citep{cui2021alphaevolve,guo2024ds,grosnit2024large,guo2025optimizing}, all of which demonstrate strong capabilities in complex scientific and engineering tasks.

Despite recent progress, current LLM agents typically follow two prevailing paradigms, each exhibiting fundamental limitations. The first approach builds specialised frameworks with fixed workflows and hard-coded reasoning, which work well for narrow tasks but lack flexibility. After deployment, such agents are static: they neither incorporate online information nor adapt to novel situations.  The second paradigm focuses on updating the LLM itself through parameter tuning of underlying LLMs -- via supervised fine-tuning or reinforcement learning -- which allows for more flexible behaviour~\citep{christianos2023pangu,shi2025pangu} but comes at a high computational cost. These approaches are inefficient for continuous adaptation and online learning, impractical for agents deployed in open-ended scenarios. This observation raises a central research challenge towards generalist agents:

\begin{center}
{\textit{How can we build LLM agents that learn continuously from a changing environment without the prohibitive cost of fine-tuning the underlying LLMs?}}
\end{center}

Inspired by human memory mechanisms, we address this challenge by proposing a memory-based learning framework that enables continual adaptation without modifying the underlying LLMs. We observe that humans' performance steadily improves because each experience is (i) encoded as an episodic trace~\citep{pritzel2017neural}, (ii) distilled into abstract rules during sleep-dependent consolidation~\citep{squire2015memory}, (iii) selectively reinforced by dopamine-driven credit assignment~\citep{glimcher2011understanding}, and (iv) retrieved through case- or analogy-based reasoning when similar problems arise~\citep{ashley1992case}. Thus, instead of fine-tuning the base model, LLM agents leverage an external memory to store past trajectories -- including successes and failures labels -- and draw from similar past experiences to guide decision making. This approach aligns with the principles of case-based reasoning (CBR)~\citep{aamodt1994case,guo2024ds,guo2025optimizing}, a psychologically grounded learning strategy supported by evidence that humans often solve problems by recalling analogous past situations~\citep{anderson2013architecture, ross1989some}. For example, in a deep research scenario, deep research agents that have previously succeeded on a web-based task can leverage their experience to solve never-seen and structurally similar tasks~\citep{wiratunga2024cbr}. Our method offers a novel path to continual learning for deep research agents -- efficient, generalizable, and inspired by how humans learn.

To this end, we introduce \emph{Memento}, a non-parametric, learn-on-the-fly framework for CBR~\citep{smyth2001similarity,hatalis2025review}, instantiated as a planner–executor architecture grounded in a memory-based Markov Decision Process (MDP). \emph{Memento} comprises three principal components: (i) a planner, (ii) a tool‑enabled executor, and (iii) a growing \emph{Case Bank} that stores past trajectories as episodic memory. Instead of relying solely on the LLM’s parametric memory, which is fixed after training, online case-based reasoning in \emph{Memento} is implemented by storing rich episodic traces.

Our experiments are conducted on $4$ benchmarks, where GAIA~\citep{mialon2023gaia} for long-horizon tool use, DeepResearcher~\citep{zheng2025deepresearcher} for real-time web research, SimpleQA~\citep{wei2024measuring} for factual precision, and HLE~\citep{phan2025humanity} for long-tail academic reasoning. We use a planner–executor architecture with GPT-$4.1$ as the planner and o$4$-mini as the default executor (o$3$ for GAIA), instrumented with tools, namely \emph{Memento}. We achieve {top-1} on GAIA validation ($87.88\%$ Pass@$3$) and $79.40\%$ on the private test leaderboard, and it reaches $66.6\%$ F$1$ and $80.4\%$ PM on the DeepResearcher dataset, outperforming the state-of-the-art training-based system, while case-based memory adds $4.7$ to $9.6$ absolute points on out-of-distribution tasks and yields $95.0\%$ PM on SimpleQA. To our knowledge, we are the first to cast case-based continual learning for LLM agents, achieving the top-level performance on the GAIA benchmark, thereby providing a principled framework for continual adaptation of Deep Research agents.  

\section{Related Work}

We first review methods that equip LLMs with continual-learning capabilities. Then, we discuss approaches that augment agents with external tools and multi-agent coordination. Lastly, we introduce agent memory mechanisms, characterising design choices in representation, retrieval, and decay, and their implications for continual learning.

\subsection{Continual-learning in LLM Agent Systems}

Continual-learning strategies for LLM agents can be categorised into two camps. \textbf{Parametric approaches}~\citep{zhu2025ai,zhu2025deepreview} update the LLM through post-training (e.g., Reinforcement Learning~\citep{wang2025otc}) or supervised fine-tuning (e.g., START~\citep{li2025start}), achieving high task fidelity at the expense of considerable compute, data, and the danger of catastrophic forgetting~\citep{li2024revisiting}. It is often assumed that achieving the capability to solve complex reasoning problems requires substantial changes to the model’s parameters, and therefore, full fine-tuning is widely applied during RL~\citep{liu2025understanding}. However, when tackling long-horizon, complex tasks~\citep{mialon2023gaia, phan2025humanity}, LLM agent systems must spend substantial time rolling out trajectories to gather training data, and they additionally depend on large volumes of human-annotated questions. Differently, \textbf{non-parametric approaches} freeze the LLM and attach an external memory to optimise the prompt construction process.  Human intelligence relies heavily on memory systems, especially episodic memory, which supports learning from both successes and failures~\citep{baddeley1983working}. Cognitive science suggests that such memories are segmented and selectively replayed to inform future decisions~\citep{anderson1997act,khosla2023survey,fountas2024human}. This inspired early AI paradigms like Case-Based Reasoning (CBR)~\citep{francis1993utility}. While modern Retrieval-Augmented Generation (RAG) systems~\citep{lewis2020retrieval} share surface similarities with CBR, they typically query static document corpora and lack mechanisms for continual adaptation~\citep{gao2023retrieval}. 

\subsection{Tool-augmented LLM}

Language agents increasingly incorporate external tools to overcome context limitations and computational bottlenecks. Prompt-based methods, including WebGPT~\citep{nakano2021webgpt}, embed tool calls directly in the generation trace. However, tackling long-horizon tasks often requires multi-hop tool calls. Therefore, recent works propose multi-agent pipelines, such as AutoGen~\citep{wu2023autogen}, OWL~\citep{owl2025} and DeerFlow~\citep{bytedance2025deerflow} to coordinate specialised agents via dialogue. To address long-horizon decision-making in dynamic, multi-turn interactions with external tool environments, Agentic Reinforcement Learning (Agentic RL) has emerged as a promising training paradigm. This approach shifts LLM training from static task-solving (e.g., math or code) to dynamic, agent–environment reasoning. Supervised Fine-tuning methods, including Toolformer~\citep{schick2023toolformer}, API-Bench~\citep{li2023api}, and GRPO-based optimisation~\citep{wang2025otc,qian2025toolrl,feng2025retool} teach models when and how to invoke APIs, but require costly retraining and often assume a fixed, small toolset (e.g., Code and Search). However, without explicit planning, deciding when and which tools to invoke remains a major bottleneck for long-horizon tasks. We model planning as a stateful MDP with explicit memory for past cases. By bringing case-based reasoning into planning, the executor is steered toward strategic tool calls and consistently strong performance.

\subsection{Agent Memory Mechanism}

Recent work has centred on endowing LLM agents with explicit memory structures. A growing body of work~\citep{owl2025, openmanus2025, google2025deepresearch, bytedance2025deerflow} has shown that current LLM agents are designed for fixed environments, limiting their ability to evolve. While some efforts, such as ReAct-style agents and reflective prompting pipelines~\citep{shinn2023reflexion, yao2023} demonstrate improvement through feedback, they remain constrained by pre-defined heuristics and do not achieve true lifelong learning. DS-Agent~\citep{guo2024ds} stabilises planning by mining prior Kaggle solutions and turning them into executable pipelines. Agent-K~\citep{grosnit2024large} adds structured memory and credit assignment to reuse past work, enabling end-to-end automation of Kaggle-style workflows. Furthermore, Agent-KB~\citep{tang2025agent} and Alita~\citep{qiu2025alita} construct shared knowledge bases and optimised toolsets for agentic problem-solving. However, most systems keep adding cases without selective curation, leading to the classic swamping problem where retrieval costs outweigh utility~\citep{francis1993utility}. 

LLM agents are increasingly equipped with long-term memory that grows and adapts over time, allowing them to accumulate knowledge, recall prior context, and adjust 
behaviour based on experience. MemoryBank~\citep{zhong2024memorybank} couples retrieval with an Ebbinghaus-style forgetting schedule so older, low-utility items decay while user-relevant facts are reinforced. Building on this idea, SAGE~\citep{liang2024self} unifies reflection with an Ebbinghaus-based memory optimiser to support continual self-refinement. Mem0~\citep{chhikara2025mem0} adopts a structured memory mechanism with explicit operations (ADD, UPDATE, DELETE, NOOP). A-MEM~\citep{xu2025mem} maintains memory via a typological network. MemInsight~\citep{salama2025meminsight} pushes further on semantics by augmenting raw memories with summaries and tags to aid retrieval. Several lines of work distil operational knowledge from interaction traces: ExpeL~\citep{zhao2024expel} collects trajectories and converts them into reusable natural-language insights and rules; AutoGuide~\citep{fu2024autoguide} compresses offline logs into concise, conditional, context-aware guidelines; and Agent Workflow Memory~\citep{wang2024agent} induces frequently used subtask sequences as auxiliary skills. Finally, Agent-KB~\citep{tang2025agent} and Alita~\citep{qiu2025alita} construct shared knowledge bases and optimised toolsets to support agentic problem solving. Differently, we formulate planning as a memory-augmented MDP and learn a neural case-selection policy over an episodic case bank via online soft Q-learning, enabling continual adaptation without fine-tuning the underlying LLM parameters.

\begin{figure}
    \centering
    \includegraphics[width=0.9\linewidth]{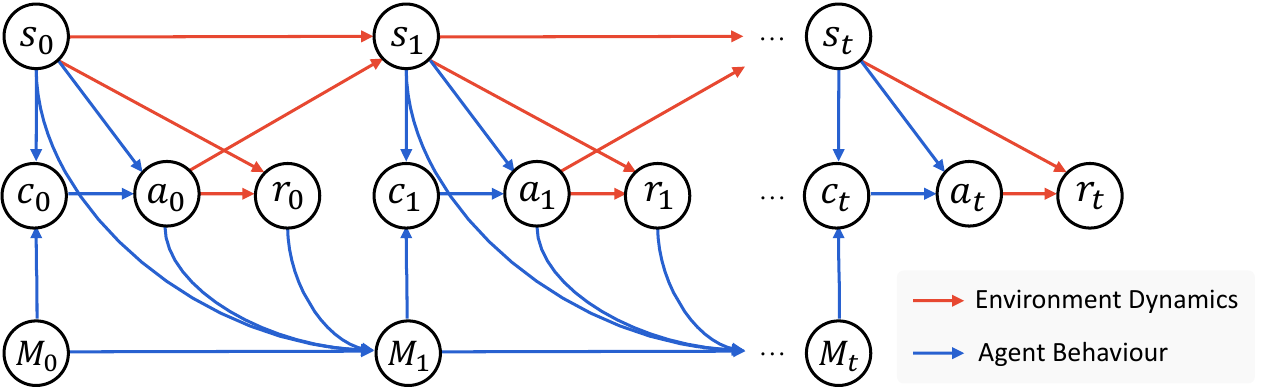}
    \caption{A graphical model of memory-based Markov Decision Process.}
    \label{fig:mmdp}
\end{figure}

\section{Methodology: Memory-Based MDP with Case-based Reasoning Policy}
\label{sec:methodology}

In this work, we integrate LLM agents with case-based reasoning, a classic problem-solving paradigm that solves new problems by learning from solutions to previously encountered similar problems. As such, LLM agents can achieve continuous improvement without parameter fine-tuning by learning from experiences stored in memory. To begin with, we model the sequential decision-making process of CBR agents as a Memory-Based Markov Decision Process (M-MDP) as below.

\begin{definition} [Memory-Based Markov Decision Process]
    A Memory-Based Markov Decision Process is a tuple $\langle \mathcal{S}, \mathcal{A}, \mathcal{P}, \mathcal{R}, \gamma, \mathcal{M} \rangle$, where $\mathcal{S}$ is the state space, 
    $\mathcal{A}$ is the action space,
    $\mathcal{P}:\mathcal{S}\times\mathcal{A}\rightarrow \Delta(\mathcal{S})$ is the transition dynamics, $\mathcal{R}:\mathcal{S}\times\mathcal{A}\rightarrow\mathbb{R}$ is the reward function, $\gamma \in [0,1)$ is the discount factor, and $\mathcal{M}=(\mathcal{S}\times\mathcal{A}\times\mathbb{R})^*$ is the memory space.
\end{definition}

The graphical model of M-MDP is illustrated in Figure \ref{fig:mmdp}. Note that the key difference from standard MDP is that we introduce a memory space as a set of past experiences. In the CBR agent setting, both state space and action space are defined as the set of all finite-length sequences over a predefined vocabulary $\mathcal{V}$.

With the M-MDP formulation, the behaviour of the CBR agent can be formally described as follows. At timestep $t$, we maintain a case bank (i.e., the memory) $M_t=\{c_i\}_{i=1}^{N_t}$, with each case $c_i$ a tuple $(s_i,a_i,r_i)$, and $N_t$ the number of cases in the current case bank. Given the current state $s_t$, the CBR agent first retrieves a case $c_t \sim \mu(\cdot \mid s_t, M_t)$, and then reuses and adapts it via the LLM, i.e., $a_t \sim p_{\text{LLM}}(\cdot \mid s_t, c_t)$. Here, $\mu$ denotes the case retrieval policy, whose implementation details will be presented later. Taking the action $a_t$, the CBR agent receives the reward $r_t=\mathcal{R}(s_t,a_t)$ and observes the next state $s_{t+1}\sim \mathcal{P}(\cdot|s_t,a_t)$. The CBR agent also retains the new case in the case bank, i.e., $M_{t+1}=M_t \cup \{(s_t,a_t,r_t)\}$. In this way, we can define the overall policy of the CBR agent as below.

\begin{definition}[Case-Based Reasoning Agent]
\label{def:cbr_agent}
A Case-Based Reasoning Agent is an agent that makes decisions based on both the current state and a finite memory of past experiences. Formally, let $s \in \mathcal{S}$ denote the current state; $M \in \mathcal{M}$ denote the current case bank, consisting of past cases $c$; $a \in \mathcal{A}$ denote the action; $\mu(c \mid s, M)$ denote a case retrieval policy, assigning a probability distribution over $M$ given the current state $s$; $p_{\text{LLM}}(a \mid s, c)$ denote the action likelihood of a large language model (LLM) conditioned on the current state $s$ and a retrieved case $c \in M$. Then, the overall policy $\pi$ of a CBR agent is defined as:
\begin{equation}
\label{policy}
  \pi(a|s,M) = \sum_{c\in M} \mu(c|s,M) p_{\text{LLM}}(a|s,c).  
\end{equation}
\end{definition} 

Overall, the trajectory $\tau$ of the CBR agent can be described as: $\tau=\{M_0,s_0,c_0,a_0,r_0,M_1,s_1,c_1,a_1,r_1,\cdots\}$. The probability of sampling the trajectory $\tau$ can be modelled as:
\begin{equation}
   p(\tau)=
\prod_{t=0}^{T-1}
\underbrace{\mu(c_t\mid s_t,M_t)}_{\textcolor[RGB]{39,97,208}{(1)~\text{Retrieve}}}\;
\underbrace{p_{\text{LLM}}(a_t\mid s_t,c_t)}_{\textcolor[RGB]{39,97,208}{(2)~\text{Reuse\&Revise}}}\;
\underbrace{\mathbb{I}[r_t=\mathcal{R}(s_t,a_t)\bigr]}_{\textcolor[RGB]{231, 72, 49}{(3)~\text{Evaluation}}}\;
\underbrace{\mathbb{I}[M_{t+1}=M_t\cup(s_t,a_t,r_t)\bigr]}_{\textcolor[RGB]{39,97,208}{(4)~\text{Retain}}}\;
\underbrace{\mathcal P(s_{t+1}\mid s_t,a_t)}_{\textcolor[RGB]{231, 72, 49}{(5)~\text{Transition}}},
\end{equation}
where $\mathbb{I}(\cdot)$ is the indicator function, assigning probability 1 if the condition holds and 0 otherwise, modelling the deterministic reward function and memory update, and $T$ denotes the maximum trajectory length. Note that the reward function and memory update can also be probabilistic in some specific cases, which we leave as future work. Among them, \textcolor[RGB]{39,97,208}{(1) Retrieve}, \textcolor[RGB]{39,97,208}{(2) Reuse and Revise}, and \textcolor[RGB]{39,97,208}{(4) Retain} describe the agent behaviour; \textcolor[RGB]{231, 72, 49}{(3) Evaluation} and \textcolor[RGB]{231, 72, 49}{(5) Transition} model the environment dynamics.

\paragraph{Soft Q-Learning for CBR Agent.} To optimise the CBR policy $\pi$ in Eq.~\eqref{policy}, we aim to learn the case retrieval policy $\mu$ with the LLM component $p_{\text{LLM}}$ fixed. In this context, the "action" of $\mu$ is to select a case $c=(s, a, r)$ from the case bank $M$. To optimise it while encouraging diversity in retrieved cases, we apply the maximum entropy RL framework \citep{haarnoja2018soft} and derive the following optimisation objective:
\begin{equation}
\label{eq:max-entropy}
J(\pi)=\mathbb{E}_{\tau\sim p}\left[\sum_{t=0}^{T-1}\left[\mathcal{R}(s_t,a_t) + \alpha \mathcal{H}\left(\mu\left(\cdot|s_t,M_t\right)\right)\right]\right],
\end{equation}
where $\mathcal{H}$ denotes the entropy, and $\alpha$ denotes the hyper-parameter of the entropy weight in the final reward.  Under this framework, the value function can be defined as:
\begin{equation}
V^{\pi}(s_t,M_t)=\sum_{c\in M_t}\mu(c|s_t,M_t)[Q^{\pi}(s_t,M_t,c)-\alpha\log\mu(c|s_t,M_t)].
\end{equation}
Also, the Q value function for taking an "action" (i.e., selecting a case), given a state, can be defined as:
\begin{equation}
Q^{\pi}(s_t,M_t,c_t)=\mathbb{E}_{a\sim p_{\text{LLM}}(\cdot|s_t,c_t), s_{t+1}\sim\mathcal{P}(\cdot|s_t,a_t)} \left[ \mathcal{R}(s_t,a_t) + \gamma V^\pi(s_{t+1},M_{t+1}) \right],
\end{equation}

where $M_{t+1}$ denotes the updated memory after $(s_t, a_t, r_t)$ is added. Let $d^{\pi}(s, M) = \sum_{t=0}^\infty \gamma^{t-1} \mathbb{P}(s_t = s, M_t = M)$ denote the discounted visitation frequency of $(s, M)$ under $\pi$. The expected value function objective is then defined as:
\begin{equation}
\begin{aligned}
J(\pi)=\mathbb{E}_{(s,M)\sim d^{\pi}}\left[V^{\pi}(s,M)\right] = \mathbb{E}_{(s,M)\sim d^{\pi}}\left[\sum_{c\in M} \mu(c|s,M)\left[ Q^{\pi}(s,M,c)- \alpha \log\mu(c|s,M) \right]\right].
\end{aligned}
\end{equation}
\iffalse
To optimise it, we apply the maximum entropy RL framework \citep{haarnoja2018soft} and derive the following optimisation objective:
\begin{equation}
\label{eq:max-entropy}
\max_\theta J_{\text{MaxEnt}}(\theta)=\mathbb{E}_{\tau\sim p_\theta}\left[\sum_{t=0}^{T-1}\left[\mathcal{R}(s_t,a_t) + \alpha \mathcal{H}\left(\mu_\theta\left(\cdot|s_t,M_t\right)\right)\right]\right],
\end{equation}
where $\mathcal{H}$ denotes the entropy, and $\alpha$ denotes the hyper-parameter of the entropy weight in the final reward. 

Following soft Q-learning \citep{haarnoja2017reinforcement}, the soft Q-function over the state $s$ and case $c$ is defined as:
\begin{equation}
Q^\pi(s,c|M)=\mathbb{E}_{a\sim p_{\text{LLM}}(\cdot|s,c), s'\sim\mathcal{P}(\cdot|s,a)} \left[ r(s,a) + \gamma V^\pi(s'|M^\prime) \right],
\end{equation}
where $M^\prime\equiv M\cup\{(s,a,r)\}$ denotes the updated case bank. The soft value function over the state $s$ is defined as:
\begin{equation}
V^\pi(s|M)=\alpha \log \sum_{c' \in M} \exp(\frac{1}{\alpha}Q^\pi(s,c'|M)).
\end{equation}
\fi
Then, we can derive the closed-form solution of the optimal retrieval policy as a softmax over the optimal Q value:
\begin{equation}
\label{eq:softmax-Q}
    \mu^*(c|s,M) = \frac{\exp(Q^*(s,M,c)/\alpha)}{\sum_{c' \in M} \exp(Q^*(s,M,c')/\alpha)}.
\end{equation}
The detailed derivation can be found in Appendix~\ref{app:der}. In this way, we can derive the optimal retrieval policy by learning the Q-function $Q$, which can be achieved by the temporal difference (TD) learning in soft Q-learning \citep{haarnoja2017reinforcement} as:
\begin{equation}
\label{eq:TD}
    Q(s_t,M_t,c_t) \leftarrow Q(s_t,M_t,c_t) + \eta \left[r_t+ \gamma \alpha \log \sum_{c' \in M_{t+1}}\exp\left(Q(s_{t+1},M_{t+1},c_{t+1})\right)-Q(s_t,M_t,c_t)\right],
\end{equation}
where $\eta$ denotes the learning rate. Next, we provide a simpler way to learn the Q-function by learning a similarity kernel over states. 

\paragraph{Enhance Q-Learning Based on State Similarity.} As in Eq.~(\ref{eq:TD}), we can learn the Q function from scratch via TD learning. However, directly learning the Q function is challenging due to complex state and case descriptions in the form of natural language.
To this end, we propose to approximate the Q value via kernel-based estimation, following episodic control~(EC) algorithms \citep{pritzel2017neural}. Specifically, we maintain an episodic memory $\mathcal{D}=\{(s,c,Q)\}$, including the state, the retrieved case, and the Q value of each interaction. Then, we approximate the Q function via a kernel network $k_\theta(\cdot,\cdot)$, parametrised by $\theta$:
\begin{equation}
\label{eq:EC}
    Q_{\text{EC}}(s,M,c;\theta) = \sum_{(s',c',Q') \in \mathcal{D}_c} \frac{k_\theta(s,s') Q'}{\sum_{(\hat{s},\hat{c},\hat{Q}) \in \mathcal{D}_c} k_\theta(s,\hat{s})},
\end{equation}
where $\mathcal D_c=\{(s_i,c_i,Q_i)\in\mathcal D:\,c_i=c\}$ denotes the past interactions stored in the episodic memory $\mathcal{D}$ with the same retrieved case $c$. By substituting Eq.~(\ref{eq:EC}) in Eq.~(\ref{eq:TD}), we can learn the Q function by optimising the kernel parameter $\theta$ via TD learning, i.e.,
\begin{equation}
\label{eq:final-td}
    \mathcal{L}(\theta) = \mathbb{E}_{(s,c,r,s',M,M')}\left[\left( Q_{\text{EC}}(s,M,c;\theta) - [r + \gamma \alpha \log \sum_{c' \in M'}\exp(Q_{\text{EC}}(s',M',c';\bar{\theta}))] \right)^2\right],
\end{equation}
where $\bar{\theta}$ denotes the target kernel network, $s^\prime$ denotes the next state and $M^\prime={M}\cup 
\{c\}$ denotes the updated case bank. More specifically, we provide the gradient of the TD learning loss with respective of $\theta$ as: 
\begin{equation}
\label{eq:final_grad}
\triangledown_\theta \mathcal L(\theta)
=2\,\mathbb E_{(s,c,r,s',M,M')}\!\left[(f_\theta(s,c)-y)\sum_{i\in\mathcal D_c} w_i(s,c;\theta)\,(Q_i-f_\theta(s,c))\,\triangledown_\theta\log k_\theta(s,s_i)\right],
\end{equation}
where $w_i=\frac{k_\theta(s,s_i)}{\sum_{s_j\in\mathcal D_c}k_\theta(s,s_j)}$, $f_\theta(s,c)=\sum_{(s_i,Q_i)\in\mathcal D_c}w_iQ_i$, and $y=r+\gamma\alpha\log\sum_{c'\in M'}\exp\!\big(f_{\bar\theta}(s',c')\big)$.

\begin{algorithm}[t]
\caption{Fine-tuning CBR agent with soft Q-learning and state similarity}
\label{alg:cbr-sql-ss}
\begin{algorithmic}[1]
\Require Kernel network parameters $\theta$, LLM policy $p_{\text{LLM}}$, entropy weight $\alpha$, discount factor $\gamma$,\, learning rate $\eta$,  target‐network update period $K$, averaging weight $\beta$, initial case bank $M_0=\emptyset$, initial episodic memory $\mathcal{D}=\emptyset$ and initial replay buffer $\mathcal{B}=\emptyset$
\State Initialize target retrieval network $\bar\theta\leftarrow \theta$
\For{timestep $t=0,1,2,\dots$}
\State \textbf{Retrieve}: Sample case $c_t\sim\mu_\theta(\cdot\mid s_t,M_t)$ \Comment{Memory Reading, following Eq. (\ref{eq:softmax-Q}) and Eq. (\ref{eq:EC})}
\State \textbf{Reuse \& Revise}: Sample action $a_t\sim p_{\text{LLM}}(\cdot\mid s_t,c_t)$
\State Execute $a_t$ and observe reward $r_t$ and next state $s_{t+1}$
\State \textbf{Retain}: $M_{t+1}=M_t\cup\{(s_t,a_t,r_t)\}$
\State Store transition $\bigl(s_t,c_t,r_t,s_{t+1},M_{t+1}\bigr)$ in $\mathcal{B}$
\State Append Episodic Memory $\mathcal{D} \leftarrow \mathcal{D} \cup \{(s_t,c_t,Q_t)\}$ \Comment{Memory Writing}
\State Sample mini-batch $\{(s_i,c_i,r_i,s_i',M_i')\}\sim\mathcal{B}$
\State $\theta\leftarrow\theta-\eta\,\triangledown_\theta\mathcal{L}_i$ \Comment{Following Eq.~\eqref{eq:final-td}}
\If{$t$ mod $K=0$} \Comment{Update target network}
    \State $\bar\theta\leftarrow \beta \bar\theta + (1-\beta)\theta$
\EndIf
\EndFor
\end{algorithmic}
\end{algorithm}

\section{Implementation: Deep Research Agent}
\label{sec:implementation}
\begin{figure}[t]
\centering
\includegraphics[width=0.9\linewidth]{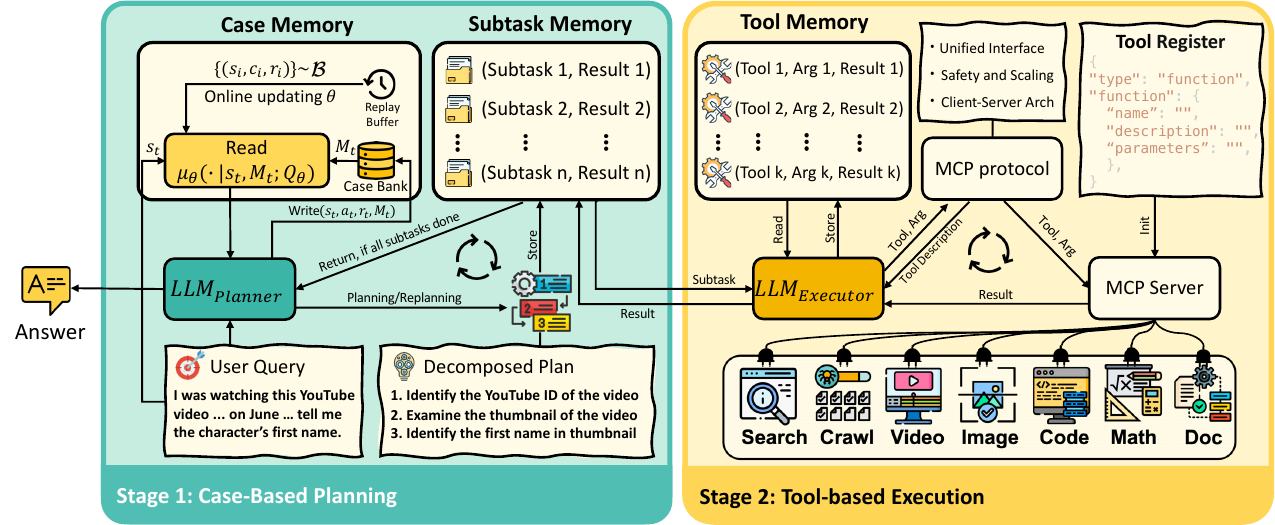}
\caption{The architecture of \emph{Memento} with parametric memory. \emph{Memento} is instantiated as a planner–executor framework alternating between Case‑Based Planning~(Stage 1) and Tool-Based Execution~(Stage 2). The planner is an LLM-based CBR agent enhanced by a Case Memory module that supports both Write, which records new cases and online refines the Q-function, and Read, which retrieves cases via the learned retrieval policy for adaptive case selection. The executor is an LLM-based MCP client that invokes external tools hosted on the MCP servers through the MCP protocol.}
\label{fig:fly}
\end{figure}

We implement stateful prompt engineering via M-MDP methodology (\S~\ref{sec:methodology}) in Deep Research scenarios~\citep{huang2025deep}, where agents must solve complex, long-horizon tasks by iteratively interacting with their environment, invoking external tools, retrieving information from external sources, and processing heterogeneous data for dynamic reasoning. As illustrated in Figure~\ref{fig:fly}, \emph{Memento} alternates between two core stages: Case-Based Planning and Tool-Based Execution. 

\subsection{Framework}

To address the challenges of long-horizon reasoning, \emph{Memento} follows the plan-and-act paradigm~\citep{erdogan2025plan}, where the planner and executor operate in an alternating loop to iteratively advance task completion. For effective coordination, \emph{Memento} integrates three memory modules:Case Memory (vectorised storage of prior cases for high-level planning), Subtask Memory (text-based storage of active subtasks and their results), and Tool Memory (text-based logs of tool interactions for each subtask).

In the planning stage, Planner, instantiated as an LLM-driven CBR agent, receives the task instruction and queries the case memory for relevant case triplets ${(s_i, a_i, r_i)}_{i=1}^K$, where $s_i$ is the task, $a_i$ is the plan, $r_i$ indicates success, and $K$ is the retrieval count. This process is supported by a Case Memory module, which retrieves relevant experiences from a case bank through either a similarity-based retriever or online-updating Q-function, thus enabling the planner to leverage both parametric and non-parametric memory as priors. The retrieved cases are concatenated with the current task instruction to form the prompt, guiding the LLM to generate a plan for each subtask. Once the initial task is decomposed, a Subtask Memory module orchestrates the interaction between the planner and executor, recording generated subtasks and their execution outcomes. After each iteration, the planner uses the accumulated execution history to assess task completion. If the task is unfinished, the planner replans based on updated context; otherwise, the final result is returned, and the case memory is updated with new experiences only upon task completion. 

The execution stage is managed by an Executor, powered by a general-purpose LLM, which is responsible for executing each subtask as an autonomous episode~\citep{sumers2023cognitive} using the MCP protocol. Unlike prior agents~\citep{zheng2025deepresearcher,weng2025cycleresearcher}, \emph{Memento}'s executor supports rich reasoning and flexible tool composition. For each subtask, the executor consults the tool memory, determines the appropriate tool invocation, and updates the results.which operates as a Model Context Protocol (MCP)\footnote{\url{https://github.com/modelcontextprotocol}} client. The executor reads pending subtasks from the subtask memory, accesses relevant history from a Tool Memory (scoped per subtask), and determines whether to invoke an external tool or return a result. MCP serves as a standardised, model-agnostic interface, enabling flexible coordination with diverse external tools and data sources. By unifying access under a single protocol layer, \emph{Memento} can seamlessly integrate dynamic reasoning and compositional tool use across multiple domains.

\subsection{Case Memory Management}
\label{sec:mem}

The case memory is an online-growing case bank $M_t$ operated with Write and Read operations, available in non-parametric and parametric variants. In the non-parametric setting, Write simply appends $(s_t,a_t,r_t)$, and Read retrieves cases by similarity for computational efficiency. In the parametric setting, Write further online updates a Q-function to shape the retrieval distribution, while Read is driven by the learned Q-function, thereby realising adaptive case selection. More details are provided in Appendix~\ref{app:mem_jus}.

\textbf{Memory Storage.} Following the CBR agent in Definition \ref{def:cbr_agent}, the Write operation appends each historical case $(s_t, a_t, r_t)$ to the case bank $M_t$, after each time step $t$:
\begin{equation}
\label{eq:write}
\text{Write}(s_t,a_t,r_t,M_t)=M_{t+1} =M_t \cup \left\{ \left( s_t,a_t,r_t\right)\right\}.
\end{equation}
In this process, the state $s_t$ is encoded using a frozen text encoder, while the action $a_t$ and reward $r_t$ are preserved in their original forms, as only the state representation requires vectorisation for subsequent retrieval operations. This Write operation is continuously performed throughout the agent’s execution, allowing the case bank to grow into a comprehensive and transferable repository of experiences incrementally. By accumulating both successes and failures, the memory not only enables retrospective analysis for informed avoidance of past mistakes but also provides successful trajectories that prospectively guide future planning.

\textbf{Non-Parametric Memory Retrieval.} A cornerstone of \emph{Memento} is its dynamically evolving Case Bank, which underpins its continual learning capability. At each planning step, this non-parametric memory module receives the task instruction and then retrieves relevant cases, comprising a mixture of successful and failed cases. This CBR method mirrors human analogical learning, where previously encountered outcomes shape decision-making~\citep{aamodt1994case}. Specifically, we retrieve the $\mathrm{K}$ nearest past cases from the case bank by computing the semantic similarity between the current state and past states. This design follows the mainstream CBR paradigm, which assumes that similar problems should have similar solutions~\citep{wiratunga2024cbr,guo2025optimizing}, thereby allowing the agent to prioritise cases whose historical contexts are most aligned with the current task. Formally, the Read operator of the non-parametric memory is defined as:
\begin{equation}
\label{eq:read_np}
\text{Read}_{\text{NP}}(s_t,M_t) = \mathop{\mathrm{TopK}}_{(s_i,a_i,r_i)\,\in\, M_t} \operatorname{sim}\big( \operatorname{enc}(s_t),\, \operatorname{enc}(s_i) \big),
\end{equation}
where $s_t$ and $M_t$ denote the query and case bank at time step $t$, respectively. Here, $\operatorname{enc}(\cdot)$ represents the pretrained textual
encoder and $\operatorname{sim}(\cdot)$ denotes the cosine similarity function.

\textbf{Parametric Memory Retrieval.} 
To empower the agent to selectively leverage high-utility cases to augment planning from past experiences, we design a differential memory mechanism in \emph{Memento} via a parametric Q-function. When writing new cases to the case bank, the parametric method, in contrast to the non-parametric approach that merely appends the tuple as in Eq.~\eqref{eq:write}, concurrently updates the Q-function online. Meanwhile, with CBR applied only for planning in \emph{Memento}, the CBR planner can be simplified to a single-step setting instead of a multi-step M-MDP. This single-step setting collapses the TD target in Eq.~\eqref{eq:final-td}  to the immediate reward, thereby simplifying the learning objective. Without bootstrapping, the updating reduces to a supervised learning paradigm, which avoids non-stationary targets. Therefore, we can train a parametric Q-function $Q(s,c;\theta)$ end-to-end, dispensing with the kernel-based estimation in Eq.~\eqref{eq:EC}. Accordingly, the single-step Q-learning loss can be formulated as:
\begin{equation}
\label{eq:dr_loss_mse}
    \mathcal{L}(\theta) = \mathbb{E}_{(s,c,r)}\left[\left( Q(s,c;\theta) - r \right)^2\right],
\end{equation}
where the tuple $\{(s,c,r)\}$ is stored in the replay buffer $\mathcal{B}$  and $Q$ is implemented as a neural network. Noting that the reward signal in deep research tasks is binary~($r\in\{0,1\}$), we replace the Mean Squared Error~(MSE) objective with a cross-entropy~(CE) loss, since MSE loss suffers from vanishing gradients near 0/1, whereas CE loss provides more numerically stable signals. Thus, we reformulate the training objective as a binary classification loss:
\begin{equation}
\label{eq:dr_loss_ce}
    \mathcal{L}(\theta) = \mathbb{E}_{(s,c,r)}\left[
        -\, r \log Q(s,c;\theta) 
        - (1-r) \log \big(1 - Q(s,c;\theta)\big)
    \right],
\end{equation}
where $Q$ can be seen as a normalised value representing the probability $p(r=1|s,c;\theta)$, i.e., the likelihood that the retrieved case $c$ is a good reference for the current state $s$ given the case bank $M$. Unlike the non-parametric approach that only preserves new cases, the parametric memory also refines the Q-function during Write, enabling each update to both record a new case and update the overall Q-value landscape.

During retrieval, the learned Q-function is used to compute the retrieval policy distribution via Eq.~\eqref{eq:softmax-Q}, from which cases can be sampled. To reduce the randomness of case selection and enhance the interpretability of the agent’s decision process, the Read operation of the parametric memory applies a $\textrm{TopK}$ operator to select the $\textrm{K}$ cases with the highest Q-values, which are used as planning references:

\begin{equation}
\label{eq:read_ec}
    % \text{Read}_{\text{EC}}(s_t,M_t) = \mathop{\mathrm{TopK}}_{c_i\,\in\, M_t}\frac{\exp(Q_{\text{EC}}(s_t,c_i)/\alpha)}{\sum_{c' \in M} \exp(Q_{\text{EC}}(s_t,c')/\alpha)},
    \text{Read}_{\text{P}}(s_t,M_t) = \mathop{\mathrm{TopK}}_{c_i\,\in\, M_t} Q(s_t,c_i;\theta).
\end{equation}
By continually updating the Q-function with new samples, the parametric memory module learns to capture the latent patterns between states and cases, thereby producing a closer approximation to the underlying distribution of the case retrieval policy $\mu^*$.

\subsection{Tool Usage}
Besides the inherent requirement for long task execution sequences and multi-turn interactions, deep research tasks also place stringent demands on the atomic actions, which require the agent to be able to acquire external information and subsequently process, integrate, and analyse it. 
Thus, we design a suite of tools for \emph{Memento} accessible via the MCP protocol, comprising modules for information retrieval such as search engines and web crawlers, as well as components for processing and analysing multimodal information, including video and image data, and files in various formats.

\textbf{External Information Acquisition.} To support open-ended tasks requiring access to up-to-date external knowledge (e.g., GAIA, BrowseComp), we design a search toolkit that integrates both retrieval and content acquisition capabilities. Specifically, we employ searxng~\footnote{\url{https://github.com/searxng/searxng-docker}}, a self-hosted metasearch engine that aggregates results from multiple sources such as Google~\footnote{\url{https://www.google.com/}}, Bing~\footnote{\url{https://www.bing.com/}}, Duckduckgo~\footnote{\url{https://duckduckgo.com/}}, and Brave~\footnote{\url{https://brave.com/}}. Retrieved candidates are then re-ranked based on semantic similarity to the query context, ensuring relevance and precision. To supplement this, we incorporate Crawl4AI~\footnote{\url{https://github.com/unclecode/crawl4ai}} to fetch and parse the full web content of selected results when deeper understanding is required by the executor. In other words, the search tool functions as a coarse filter based on keyword matching in the user query, while the crawler serves as a fine-grained mechanism to extract detailed information from the retrieved sources when necessary.

\textbf{Multimodal Heterogeneous Information Processing.} To support downstream reasoning over heterogeneous data sources, we implemented a versatile and fine-grained document processing toolkit that automatically extracts information from a broad spectrum of file types and modalities. For example, images are captioned using a vision-language model~(VLM); audio is transcribed via automated speech recognition; PowerPoint files are parsed slide-by-slide with embedded image descriptions; spreadsheets are converted to a readable row-wise layout; archives are unpacked; plain text and code files are read directly; JSON and XML are parsed into structured objects; Word documents are translated into Markdown; and videos receive natural-language summaries from VLMs. For PDFs or unsupported formats, a fallback extraction via Chunkr AI~\footnote{\url{https://chunkr.ai/}} or plain-text parsing is used. This toolkit offers a unified interface for accessing and interpreting content across diverse file types and modalities, streamlining the handling of heterogeneous data in real-world scenarios.

\textbf{Reasoning.} The reasoning and analysis toolkit integrates code execution and mathematical computation to support robust, automated analysis within the \emph{Memento} framework. The Code tool provides a sandboxed environment for writing, running, and managing code within a unified workspace. Users can create files, execute shell or Python commands, and inspect outputs -- all within a persistent task directory. Python scripts are validated against a security whitelist to ensure safe execution, supporting commonly used libraries such as numpy, pandas, and torch. The workspace maintains state across steps, enabling iterative development. This agent is crucial for solving data analysis, automation, or dynamic code generation tasks. Complementing this, the Math tool handles fundamental arithmetic operations.

\section{Experiments}
\label{sec:experiments}

In this paper, we investigate the Deep Research agent, which necessitates tool use and supports multiple rounds of interaction with external, real-world environments. To comprehensively evaluate the agent's capabilities, we select four datasets, each representing a distinct aspect of the research challenge: (i) long-horizon tool use and planning (GAIA)~\citep{mialon2023gaia}, (ii) real-time web-based research (DeepResearcher)~\citep{zheng2025deepresearcher}, (iii) concise factual accuracy (SimpleQA)~\citep{wei2024measuring}, and (iv) exploration at the frontier of human knowledge (HLE)~\citep{phan2025humanity}.

\subsection{Datasets}

To evaluate the general-purpose reasoning capabilities of \emph{Memento}, we adopt the GAIA benchmark~\citep{mialon2023gaia}, which comprises 450 non-trivial questions with unambiguous answers -- 300 in the test set and 150 in the validation set. Each question requires varying levels of tool use and autonomous planning, and the dataset is stratified into three difficulty levels: Level 1: Requires approximately 5 steps using a single tool; Level 2: Requires 5–10 steps involving multiple tools; Level 3: Involves up to 50 steps with no restrictions on the number or type of tools. Each level includes a public validation split and a private test split with hidden ground-truth answers and metadata.

We further evaluate \emph{Memento} on broader benchmarks compiled in DeepResearcher~\citep{zheng2025deepresearcher}, which draws from seven open-domain QA datasets: Natural Questions (NQ)~\citep{kwiatkowski2019natural}, TriviaQA (TQ)~\citep{joshi2017triviaqa}, HotpotQA~\citep{yang2018hotpotqa}, 2Wiki~\citep{ho2020constructing}, MusiQue~\citep{trivedi2022musique}, Bamboogle~\citep{press2022measuring}, and PopQA~\citep{mallen2022not}. Each dataset contributes 512 examples, except Bamboogle, which provides 125 high-quality samples curated to minimise contamination and emphasise web-based synthesis.

Additionally, we include two challenging benchmarks: 1) SimpleQA~\citep{wei2024measuring}, consisting of 4,330 fact-seeking questions, focuses on factual accuracy. 2) Humanity’s Last Exam (HLE)~\citep{phan2025humanity}, with 2,500 questions across diverse academic subjects, assesses the limits of broad-domain reasoning.

\begin{table}[t]
\centering
\resizebox{\linewidth}{!}{
\begin{tabular}{l 
  >{\color{black}}c >{\color{black}}c
  >{\color{black}}c >{\color{black}}c
  >{\color{black}}c >{\color{black}}c
  >{\color{black}}c >{\color{black}}c
  >{\color{black}}c >{\color{black}}c
  >{\color{black}}c >{\color{black}}c
  >{\color{black}}c >{\color{black}}c
  >{\color{black}}c >{\color{black}}c}
\toprule
Method 
  & \multicolumn{2}{c}{NQ} 
  & \multicolumn{2}{c}{TQ} 
  & \multicolumn{2}{c}{HotpotQA} 
  & \multicolumn{2}{c}{2Wiki}
  & \multicolumn{2}{c}{Musique}
  & \multicolumn{2}{c}{Bamboogle}
  & \multicolumn{2}{c}{PopQA} 
  & \multicolumn{2}{c}{Avg}\\
\cmidrule(lr){2-3} \cmidrule(lr){4-5} \cmidrule(lr){6-7} \cmidrule(lr){8-9}
\cmidrule(lr){10-11} \cmidrule(lr){12-13} \cmidrule(lr){14-15} \cmidrule(lr){16-17}
  & \textcolor{black}{F1} & \textcolor{black}{PM} & \textcolor{black}{F1} & \textcolor{black}{PM} & \textcolor{black}{F1} & \textcolor{black}{PM} & \textcolor{black}{F1} & \textcolor{black}{PM} & \textcolor{black}{F1} & \textcolor{black}{PM} & \textcolor{black}{F1} & \textcolor{black}{PM} & \textcolor{black}{F1} & \textcolor{black}{PM} & \textcolor{black}{F1} & \textcolor{black}{PM} \\
\midrule
\multicolumn{17}{l}{\textbf{Prompt Based}} \\ \midrule
CoT               & 19.8 & 32.0 & 45.6 & 48.2 & 24.4 & 27.9 & 26.4 & 27.3 &  8.5 &  7.4 & 22.1 & 21.6 & 17.0 & 15.0 & 23.6 & 26.1 \\
CoT + RAG         & 42.0 & 59.6 & 68.9 & 75.8 & 37.1 & 43.8 & 24.4 & 24.8 & 10.0 & 10.0 & 25.4 & 27.2 & 46.9 & 48.8 & 37.7 & 43.2 \\
Search-o1 (Web)~\citep{li2025search}         & 32.4 & 55.1 & 58.9 & 69.5 & 33.0 & 42.4 & 30.9 & 37.7 & 14.7 & 19.7 & 46.6 & 53.6 & 38.3 & 43.4 & 35.2 & 45.0 \\
\midrule
\multicolumn{17}{l}{\textbf{Training Based}} \\\midrule
Search-r1-base~\citep{jin2025search}      & 45.4 & 60.0 & 71.9 & 76.2 & 55.9 & 63.0 & 44.6 & 47.9 & 26.7 & 27.5 & 56.5 & 57.6 & 43.2 & 47.0 & 48.3 & 53.8 \\
Search-r1-instruct~\citep{jin2025search}  & 33.1 & 49.6 & 44.7 & 49.2 & 45.7 & 52.5 & 43.4 & 48.8 & 26.5 & 28.3 & 45.0 & 47.2 & 43.0 & 44.5 & 39.6 & 45.6 \\
R1-Searcher~\citep{song2025r1}         & 35.4 & 52.3 & 73.1 & 79.1 & 44.8 & 53.1 & 59.4 & 65.8 & 22.8 & 25.6 & 64.8 & 65.6 & 42.7 & 43.4 & 47.1 & 53.7 \\
DeepResearcher~\citep{zheng2025deepresearcher}     & 39.6 & 61.9 & 78.4 & 85.0 & 52.8 & 64.3 & 59.7 & 66.6 & 27.1 & 29.3 & 71.0 & 72.8 & 48.5 & 52.7 & 51.8 & 60.5 \\
\midrule
% \multicolumn{17}{l}{\textbf{Executor-only Based}} \\\midrule
% Offline Executor (o4-mini) & \textcolor{black}{39.7} & \textcolor{black}{70.1} & \textcolor{black}{75.8} & \textcolor{black}{89.1}
%  & \textcolor{black}{50.7} & \textcolor{black}{67.2} & \textcolor{black}{44.8} & \textcolor{black}{56.0}
%  & \textcolor{black}{26.9} & \textcolor{black}{35.7} & \textcolor{black}{76.0} & \textcolor{black}{84.0}
%  & \textcolor{black}{48.0} & \textcolor{black}{53.5} & \textcolor{black}{48.8} & \textcolor{black}{62.8} \\
 
% Online Executor (o4-mini) & \textcolor{black}{23.3} & \textcolor{black}{55.3} 
% & \textcolor{black}{41.9} & \textcolor{black}{80.7} 
% & \textcolor{black}{34.4} & \textcolor{black}{67.8} 
% & \textcolor{black}{33.6} & \textcolor{black}{66.2} 
% & \textcolor{black}{23.0} & \textcolor{black}{39.7} 
% & \textcolor{black}{45.8} & \textcolor{black}{77.6} 
% & \textcolor{black}{24.9} & \textcolor{black}{50.2} & \textcolor{black}{30.8} & \textcolor{black}{60.7} \\\midrule

\multicolumn{17}{l}{\textbf{Ours}} \\\midrule
% Memento  w/o CBR (GPT-4.1 + o4-mini) & \textcolor{black}{39.5} & \textcolor{black}{67.8} & \textcolor{black}{81.8} & \textcolor{black}{89.1}
% & \textcolor{black}{62.0} & \textcolor{black}{76.0} & \textcolor{black}{78.3} & \textcolor{black}{90.0}
% & \textcolor{black}{35.6} & \textcolor{black}{43.8} & \textcolor{black}{77.5} & \textcolor{black}{83.2}
% & \textcolor{black}{57.9} & \textcolor{black}{63.9} & \textcolor{black}{59.9} & \textcolor{black}{72.2} \\
Memento (GPT-4.1 + o4-mini) & \textbf{\textcolor{black}{42.0}} & \textbf{\textcolor{black}{74.6}} & \textbf{\textcolor{black}{85.5}} & \textbf{\textcolor{black}{93.9}}
& \textbf{\textcolor{black}{66.5}} & \textbf{\textcolor{black}{81.6}} & \textbf{\textcolor{black}{81.4}} & \textbf{\textcolor{black}{94.1}}
& \textbf{\textcolor{black}{40.6}} & \textbf{\textcolor{black}{53.3}} & \textbf{\textcolor{black}{86.2}} & \textbf{\textcolor{black}{92.8}}
& \textbf{\textcolor{black}{64.0}} & \textbf{\textcolor{black}{72.5}} & \textbf{\textcolor{black}{66.6}} & \textbf{\textcolor{black}{80.4}} \\
\bottomrule
 \end{tabular}}
\caption{Performance comparison of prompt-based, training-based, and our approach on seven open-domain QA datasets. We report the F$1$ score and PM scores. The last two columns give the weighted average (\textit{Avg}) across all benchmarks, where Bamboogle contributes $125$ examples and every other dataset $512$ examples. The results of prompt-based and training-based methods using Qwen$2.5$ ($7$B) are referred to \emph{DeepResearcher}~\citep{zheng2025deepresearcher}.}
\label{table:deepresearcher}
\end{table}

\subsection{Evaluation Metrics} 

As each GAIA query has a single reference answer, we follow the GAIA leaderboard and use the Exact Match (EM) metric, which marks a prediction as correct only if it exactly matches the ground-truth answer after standard normalisation (lowercasing, punctuation and article removal, whitespace normalisation). The EM score reflects the percentage of perfectly matched answers. 

However, EM cannot accurately reflect the capabilities of an LLM agent, as it overlooks the diversity of expression. We use the macro-F1 score to evaluate the DeepResearcher, SimpleQA, and HLE datasets. Meanwhile, Partial Match (PM) indicates the partial semantic match scores between LLMs’ generated answers and gold answers. We utilise GPT-4o-mini as the answer evaluator to give the scores and the prompt the same as DeepResearcher~\citep{zheng2025deepresearcher}.

\subsection{Model Configurations} 

The Planner is powered by \texttt{GPT-4.1}, the Executor by \texttt{o3} for GAIA and \texttt{o4-mini} for other datasets, the image processing by \texttt{GPT-4o}, the video agent by \texttt{Gemini 2.5 Pro} and the audio agent by~\texttt{Assembly AI}. For the non-parametric CBR, we encode sentences with \texttt{SimCSE} and rank candidate cases using cosine similarity. For the parametric CBR, we initialise sentence representations with \texttt{SimCSE} and implement the Q-function as a two-layer MLP to assign the Q value. Meanwhile, the CBR planner’s state at each step often contains information inherited from previous states. To avoid redundant storage, only the state, action, and reward from the final step of each trajectory are written to memory, ensuring that the case bank remains both compact and informative.

The Offline Executor setting refers to one static executor, removing the planner, case memory, and all external tools, so it reflects only raw parametric knowledge from LLMs. The Online Executor starts from that stripped-down baseline but reconnects the same executor to live search and other MCP tools, reflecting the value of real-time retrieval and tool execution. \emph{Memento} (w/o CBR) keeps episodic memory disabled, allowing us to measure the extra gain delivered specifically by case-based reasoning.

\subsection{Experimental Results}

\begin{table}[t]
\centering
% \rowcolors{2}{gray!10}{white}
\resizebox{\linewidth}{!}{
\begin{tabular}{@{}llccccc@{}}
\toprule
\textbf{Agent name} & \textbf{Model family}  & \textbf{Average score (\%)} & \textbf{Level 1 (\%)} & \textbf{Level 2 (\%)} & \textbf{Level 3 (\%)} \\ 

\midrule
\multicolumn{6}{c}{\textbf{Valiadation Dataset}} \\
\midrule

\textcolor{black}{\textbf{\emph{Memento} (Pass@3)}} & \textcolor{black}{\textbf{GPT4.1, o3}} & \textcolor{black}{\textbf{87.88}} & \textcolor{black}{\textbf{96.23}} & \textcolor{black}{\textbf{90.70}} & 61.54 \\

Alita & Claude 4 Sonnet, GPT-4o & 87.27 & 88.68 & 89.53 & \textcolor{black}{\textbf{76.92}}       \\

Skywork Super Agents v1.1 & skywork-agent, Claude 3.7 Sonnet, Whisper  & 82.42  & 92.45 & 83.72 & 57.69  \\

Langfun Agent & Gemini 2.5 Pro & 79.39 & 88.68 & 80.23 & 57.69\\

AWorld  & GPT-4o, DeepSeek-V3, Claude 4, Gemini 2.5 Pro & 77.58 & 88.68 & 77.91 & 53.85 \\

Manus & - & 73.30 & 86.50 & 70.10 & 57.70 \\

% Co-sight  & claude-3.7-sonnet,gemini-2.5,gpt-4o & 71.52 & 86.79 & 72.09 & 38.46 \\

OWL-Workforce  & Claude 3.7 Sonnet & 69.09 & 84.91 & 67.44 & 42.31 \\

OpenAI DeepResearch & o3 & 67.40 & 74.30 & 69.10 & 47.60   \\

OWL-Roleplaying & GPT-4o and o3-mini & 58.18 & 81.13 & 54.65 & 23.08 \\

Open Deep Research & o1 & 55.15 & 67.92 & 53.49 & 34.62\\

\midrule
\multicolumn{6}{c}{\textbf{Test Dataset}} \\
\midrule

Su Zero Ultra & -- & \textcolor{black}{\textbf{80.40}} & \textcolor{black}{\textbf{93.55}} & 77.36 & 65.31 \\

h2oGPTe Agent v1.6.33 & Claude 3.7 Sonnet, Gemini 2.5 Pro & 79.73 & 89.25 & \textcolor{black}{\textbf{79.87}} & 61.22 \\

\textcolor{black}{\textbf{\emph{Memento}}} & \textcolor{black}{\textbf{GPT4.1, o3}} & 79.40 & 90.32 & 75.47 & \textcolor{black}{\textbf{71.43}} \\

h2oGPTe Agent v1.6.32 & Claude 3.7 Sonnet, Gemini 2.5 Pro & 79.07 & 90.32 & 77.99 & 61.22 \\

Aworld & GPT-4o, DeepSeek-V3, Claude 4 sonnet, Gemini 2.5 Pro & 77.08 & 93.55 & 76.73 & 46.94 \\

\bottomrule
\end{tabular}}
\caption{Top results on the GAIA Leaderboard as of June 26, 2025, \emph{Memento} achieves the Top-1 performance on the validation set and the test set in open-source agent frameworks.}
\label{table:1}
\end{table}

\textbf{Deep Researcher.} We include this dataset to test real-time web research, evidence retrieval, cross-page synthesis, and multi-hop reasoning. As shown in Table~\ref{table:deepresearcher}, \emph{Memento} augmented with MCP tools (e.g., search engine, browser) reaches an average $66.6 \%$ F1 across the seven DeepResearcher benchmarks, nearly doubling the $37.7 \%$ F1 of the CoT + RAG baseline.  This demonstrates that real-time, online retrieval tools can rival or even exceed carefully curated static databases.

GAIA (Validation \& Test). To assess robustness in long-horizon planning, tool orchestration, and execution, we employ the GAIA benchmark. \emph{Memento} attains the top-$1$ ranking on the validation set and $4$th place on the test set, outperforming most existing agent frameworks (Table~\ref{table:1}). Notably, it surpasses widely used open-source frameworks, including Manus~\citep{openmanus2025}, Aworld~\citep{aworld2025}, and OWL~\citep{owl2025}, on both validation and test sets.

For the GAIA validation evaluation, we initialise memory from scratch and iteratively store both successful and failed trajectories in the case bank over three iterations. Using GPT-4.1 as the planner and o3 as the executor, \emph{Memento} achieves $87.88\%$ accuracy on the validation set. For the GAIA test set, performance is based solely on the case bank accumulated during validation, yielding an accuracy of $79.40\%$. Although \emph{Memento} demonstrates strong overall performance, challenges remain for Level~3 tasks that require extended reasoning horizons and advanced tool coordination.

 \begin{figure}[t]
  \centering
  \includegraphics[width=\linewidth]{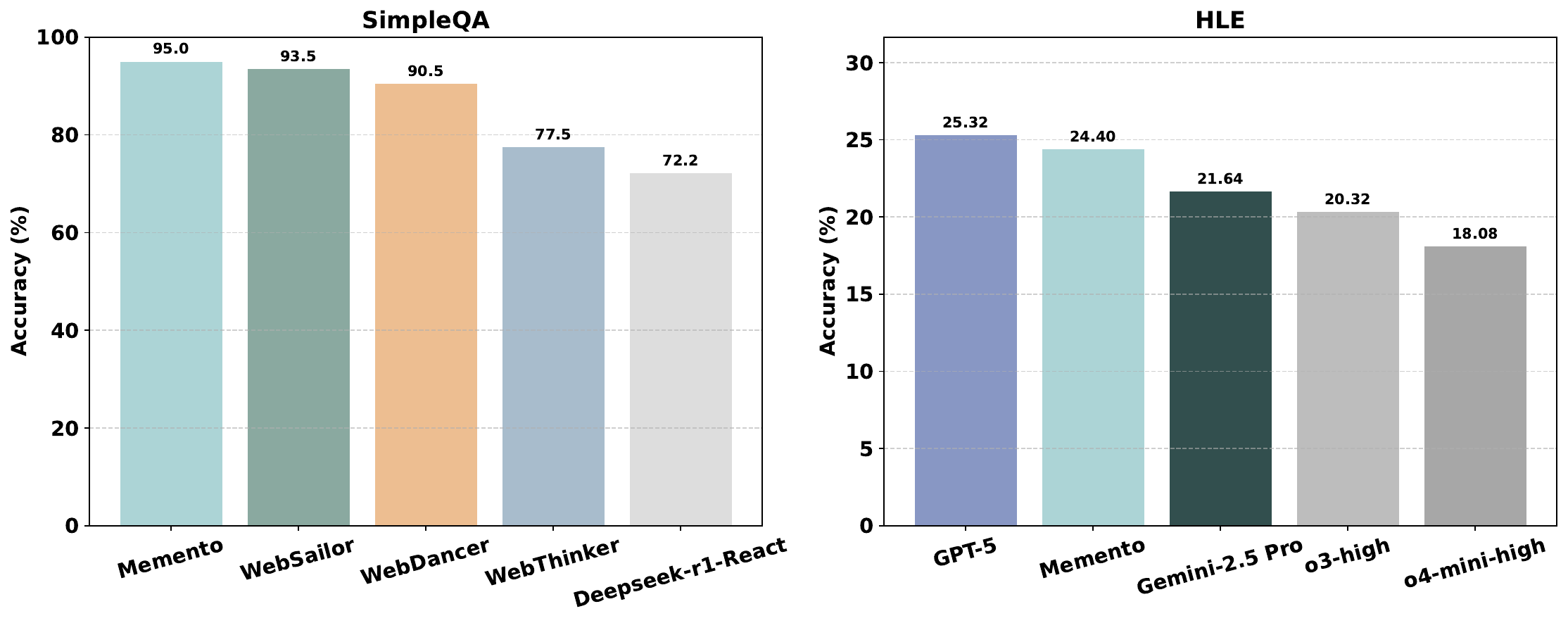}
  \caption{Performance on SimpleQA and HLE. The SimpleQA results are from WebSailor~\citep{li2025websailor}, and the HLE results are from the official website.}
  \label{fig:simpleqa-hle}
\end{figure}

\textbf{Humanity’s Last Exam (HLE).} 
To evaluate the frontier of human knowledge and the complex reasoning ability in long-tail, specialised domains, we include the HLE~\footnote{\url{https://scale.com/leaderboard/humanitys_last_exam}}. Using our planner executor architecture, with planner GPT-$4.1$ and executor o$4$-mini with tools, \emph{Memento} attains $24.4\%$ PM, ranking second overall and within $0.92$ points of GPT-$5$ at $25.32\%$, while outperforming Gemini-$2.5$-Pro at $21.64\%$, o$3$-high at $20.32\%$, and o$4$-mini-high at $18.08\%$. These results demonstrate that continual learning through CBR effectively transforms episodic experiences into reusable knowledge, offering a complementary pathway to generalisation even in long-tail domains where conventional tool usage and retrieval methods struggle.

\textbf{SimpleQA.} To evaluate \emph{Memento}’s reliability and robustness against hallucination in single-hop factual question answering, we employ the SimpleQA benchmark. As illustrated in Figure~\ref{fig:simpleqa-hle}, \emph{Memento}, implemented with a planner–executor framework (GPT-4.1 as the planner and o4-mini as the executor) augmented with tool use, achieves the highest accuracy among all baselines. Specifically, it reaches an accuracy of $95.0\%$, outperforming WebSailor ($93.5\%$), WebDancer ($90.5\%$), WebThinker ($77.5\%$), and DeepSeek-r1-React ($72.2\%$). These results demonstrate that \emph{Memento} provides strong factual reliability and substantially mitigates hallucination on straightforward single-hop queries, establishing a new state-of-the-art over prior web-agent baselines.

\subsection{Ablation Studies}

We analyse \emph{Memento}’s hyper-parameter selection, component-wise Analysis, learning curves for both parametric and non-parametric case-based reasoning, out-of-distribution performance, and token costs.

\begin{table}[t]
\centering
\resizebox{\linewidth}{!}{
\begin{tabular}{lcccccccccccccc}
\hline
\textbf{Dataset} 
& \multicolumn{2}{c}{\textbf{K=0}} 
& \multicolumn{2}{c}{\textbf{K=1}} 
& \multicolumn{2}{c}{\textbf{K=2}} 
& \multicolumn{2}{c}{\textbf{K=4}} 
& \multicolumn{2}{c}{\textbf{K=8}} 
& \multicolumn{2}{c}{\textbf{K=16}} 
& \multicolumn{2}{c}{\textbf{K=32}} \\
\cline{2-15}
& F1 & PM & F1 & PM & F1 & PM & F1 & PM & F1 & PM & F1 & PM & F1 & PM \\
\hline
NQ~\citep{kwiatkowski2019natural}        & 39.5  &67.8 & 41.1 & 74.4 & 41.3 & 72.7 & 41.9 & 73.0 & 41.7 & 73.8 & 42.1 & 73.2 & 42.2 & 75.4  \\
TQ~\citep{joshi2017triviaqa}        & 81.1 & 89.1 & 86.1 & 93.8 & 86.2 & 93.9 & 86.3 & 94.1 & 85.8 & 94.1 & 85.9 & 94.3 & 85.5 & 93.9 \\
HotpotQA~\citep{yang2018hotpotqa}  & 62.0 & 76.0 & 65.4 & 80.7 & 65.7 & 81.3 & 67.4 & 84.2 & 66.6 & 82.0 & 65.5 & 82.0 & 66.4 & 83.2  \\
2Wiki~\citep{ho2020constructing}     & 78.3 & 90.0 & 81.3 & 94.9 & 80.9 & 94.1 & 81.0 & 94.7 & 82.0 & 94.5 & 81.1 & 93.6 & 81.0 & 94.1  \\
Musique~\citep{trivedi2022musique}   & 35.6 & 43.8 & 39.8 & 50.2 & 40.1 & 52.1 & 41.6 & 51.0 & 41.0 & 52.1 & 39.6 & 51.4 & 40.3 & 50.4  \\
Bamboogle~\citep{press2022measuring} &77.5 & 83.2  & 85.9 & 91.2 & 84.1 & 91.2 & 84.7 & 90.4 & 84.9 & 91.2 & 85.2 & 92.0 & 83.0 & 87.2  \\
PopQA~\citep{mallen2022not}     & 57.9 & 63.9 & 62.6 & 70.1 & 63.4 & 71.3 & 63.6 & 70.9 & 62.7 & 69.2 & 64.1 & 70.9 & 63.2 & 69.4  \\
\hline
\textbf{Average} 
& 59.9 & 72.2 
& 63.6 & 77.9
& 63.7 & 78.1
& \textbf{64.5} & \textbf{78.5}
& 64.1 & 78.2
& 63.9 & 78.1
& 63.9 & 78.1 \\
\hline
\end{tabular}}
\caption{The performance of Memento on the DeepResearcher dataset across different numbers of cases. We use gpt-$4.1$ as the planner and o$4$-mini as the executor.}
\label{table:k-selection}
\end{table}
\subsubsection{Hyper-parameter Selection}

Increasing the number of retrieved cases in CBR raises computational cost and can introduce noise from irrelevant examples. To evaluate this, we vary $K$ in ${0, 2, 4, 8, 16, 32}$ on the DeepResearcher dataset. As shown in Table~\ref{table:k-selection}, performance improves up to $K=4$ -- yielding the highest F1 (64.5) and PM (78.5) -- but plateaus or slightly declines for larger $K$. This suggests that CBR benefits from a small, high-quality memory, unlike few-shot prompting, where more examples often help~\citep{agarwal2024many}. Careful case selection and memory curation are thus crucial for continual learning.

\begin{table}[t]
  \centering
  \footnotesize
  \begin{tabular}{lccccc}
    \toprule
    \textbf{Method} & \textbf{Iter 1} & \textbf{Iter 2} & \textbf{Iter 3} & \textbf{Iter 4} & \textbf{Iter 5} \\
    \midrule
    \multicolumn{6}{c}{Baseline} \\
    \midrule
    \emph{Memento} w/o CBR        & 78.65 & 80.93 & 82.62 & 83.53 & 84.47 \\
    \midrule
    \multicolumn{6}{c}{Case-based Continual Learning} \\
    \midrule    
    \emph{Memento} w/ Non-Parametric CBR          & 79.84 & 81.87 & 83.09 & 84.03 & 84.85 \\
    \emph{Memento}  w/ Parametric CBR   & \textbf{80.46} & \textbf{82.84} &\textbf{84.10} & \textbf{84.85} & \textbf{85.44}\\
    \bottomrule
  \end{tabular}
    \caption{Performance improvement of \emph{Memento} over five learning iterations on the DeepResearcher dataset, demonstrating the benefit of accumulating cases in the Case Bank.}
\label{table:ec}
\end{table}

% 单元格内容用斜杠分隔
\newcommand{\pair}[1]{#1} % 兼容占位
\renewcommand{\pair}[2]{#1/#2}

\begin{table*}[t]
\centering
\setlength{\tabcolsep}{4pt}
\renewcommand{\arraystretch}{1.08}

% ================== (a) HLE ==================
\begin{subtable}{\textwidth}
\centering
\footnotesize
\resizebox{\textwidth}{!}{
\begin{tabular}{l *{9}{c}}
\toprule
Model &
Humanities/SC& Math& Chemistry& Other&
Physics& Engineering& Biology/Medicine&
CS/AI& Avg\\
\midrule
Offline Executor
  & \pair{5.2}{9.6}   & \pair{7.1}{5.8}   & \pair{2.3}{7.9}   & \pair{2.9}{11.4}
  & \pair{7.6}{4.6}   & \pair{12.8}{14.0} & \pair{5.2}{17.7} & \pair{6.5}{11.9}
  & \pair{6.4}{8.7} \\
Online Executor
  & \pair{10.8}{24.9} & \pair{13.1}{16.0} & \pair{6.9}{9.3}   & \pair{14.0}{16.3}
  & \pair{7.8}{10.2}  & \pair{7.5}{5.3}   & \pair{5.7}{17.2}  & \pair{13.3}{15.4}
  & \pair{11.2}{15.8} \\
Memento w/o CBR
  & \pair{25.5}{29.2} & \pair{24.9}{16.3} & \pair{17.4}{21.1} & \pair{24.8}{24.1}
  & \pair{18.4}{10.8} & \pair{15.8}{8.8}  & \pair{10.0}{18.7} & \pair{25.4}{12.4}
  & \pair{22.2}{17.4} \\
Memento
  & \pair{\textbf{28.4}}{\textbf{33.0}} & \pair{\textbf{30.9}}{\textbf{24.2}}
  & \pair{\textbf{18.7}}{\textbf{22.7}} & \pair{\textbf{28.5}}{\textbf{32.4}}
  & \pair{\textbf{22.9}}{\textbf{19.1}} & \pair{\textbf{15.9}}{\textbf{12.1}}
  & \pair{\textbf{14.0}}{\textbf{26.1}} & \pair{\textbf{28.5}}{\textbf{18.5}}
  & \pair{\textbf{26.7}}{\textbf{24.4}} \\
\bottomrule
\end{tabular}}
\caption{HLE}
\end{subtable}

\vspace{0.6em}

% ================== (b) SimpleQA ==================
\begin{subtable}{\textwidth}
\centering
\footnotesize
\resizebox{\textwidth}{!}{
\begin{tabular}{l *{11}{c}}
\toprule
Model &
Art& Geography& Science \& Tech& Politics&
Sports& Other& TV Shows& Music&
History& Video Games& Avg\\
\midrule
Offline Executor
  & \pair{16.5}{19.6} & \pair{25.7}{31.1} & \pair{20.1}{24.7} & \pair{25.8}{24.8}
  & \pair{18.8}{19.0} & \pair{15.8}{14.9} & \pair{12.6}{13.3} & \pair{15.7}{17.6}
  & \pair{25.9}{26.6} & \pair{15.5}{13.3} & \pair{19.7}{21.5} \\
Online Executor
  & \pair{49.7}{82.9} & \pair{45.1}{82.5} & \pair{59.4}{87.7} & \pair{51.6}{86.9}
  & \pair{46.6}{90.1} & \pair{48.7}{86.4} & \pair{39.5}{84.3} & \pair{43.1}{88.5}
  & \pair{49.7}{83.8} & \pair{34.7}{78.6} & \pair{48.5}{84.8} \\
Memento w/o CBR
  & \pair{83.8}{92.2} & \pair{71.8}{84.9} & \pair{87.1}{94.1} & \pair{83.7}{90.8}
  & \pair{77.4}{86.4} & \pair{81.2}{90.7} & \pair{70.0}{81.6} & \pair{80.7}{89.1}
  & \pair{79.1}{86.1} & \pair{81.2}{88.9} & \pair{81.0}{89.7} \\
Memento
  & \pair{\textbf{86.9}}{\textbf{96.4}} & \pair{\textbf{76.6}}{\textbf{91.7}}
  & \pair{\textbf{89.3}}{\textbf{96.9}} & \pair{\textbf{87.1}}{\textbf{95.6}}
  & \pair{\textbf{82.2}}{\textbf{93.5}} & \pair{\textbf{84.2}}{\textbf{95.4}}
  & \pair{\textbf{77.3}}{\textbf{90.8}} & \pair{\textbf{83.3}}{\textbf{95.0}}
  & \pair{\textbf{85.7}}{\textbf{94.8}} & \pair{\textbf{86.3}}{\textbf{95.6}}
  & \pair{\textbf{84.7}}{\textbf{95.0}} \\
\bottomrule
\end{tabular}}
\caption{SimpleQA}
\end{subtable}

\vspace{0.6em}

% ================== (c) Open-domain QA ==================
\begin{subtable}{\textwidth}
\centering
\footnotesize
\resizebox{\textwidth}{!}{
\begin{tabular}{l *{8}{c}}
\toprule
Method &
NQ& TQ& HotpotQA& 2Wiki&
Musique& Bamboogle& PopQA& Avg\\
\midrule
Offline Executor 
  & \pair{39.7}{70.1} & \pair{75.8}{89.1} & \pair{50.7}{67.2} & \pair{44.8}{56.0}
  & \pair{26.9}{35.7} & \pair{76.0}{84.0} & \pair{48.0}{53.5} & \pair{48.8}{62.8} \\
Online Executor 
  & \pair{23.3}{55.3} & \pair{41.9}{80.7} & \pair{34.4}{67.8} & \pair{33.6}{66.2}
  & \pair{23.0}{39.7} & \pair{45.8}{77.6} & \pair{24.9}{50.2} & \pair{30.8}{60.7} \\
Memento w/o CBR 
  & \pair{39.5}{67.8} & \pair{81.8}{89.1} & \pair{62.0}{76.0} & \pair{78.3}{90.0}
  & \pair{35.6}{43.8} & \pair{77.5}{83.2} & \pair{57.9}{63.9} & \pair{59.9}{72.2} \\
Memento 
  & \pair{\textbf{42.0}}{\textbf{74.6}} & \pair{\textbf{85.5}}{\textbf{93.9}}
  & \pair{\textbf{66.5}}{\textbf{81.6}} & \pair{\textbf{81.4}}{\textbf{94.1}}
  & \pair{\textbf{40.6}}{\textbf{53.3}} & \pair{\textbf{86.2}}{\textbf{92.8}}
  & \pair{\textbf{64.0}}{\textbf{72.5}} & \pair{\textbf{66.6}}{\textbf{80.4}} \\
\bottomrule
\end{tabular}}
\caption{DeepResearcher}
\end{subtable}
\caption{Ablation results across three benchmarks. Each cell shows \textit{F1/PM}. We use gpt-$4.1$ as the planner and o$4$-mini as the executor.}
\label{tab:ablation_all}
\end{table*}

\subsubsection{Component-wise Analysis}  
From Table~\ref{tab:ablation_all}, we observe a consistent pattern across HLE, SimpleQA, and DeepResearcher. Moving from an offline executor to live tools generally reduces hallucination and increases both F$1$ and PM, though the magnitude depends on task type (SimpleQA: $+28.8$ F$1$ / $+63.3$ PM, HLE: $+4.8$ / $+7.1$), and may even hurt on open-domain data (DeepResearcher: $-18.0$ / $-2.1$). Introducing planning (\emph{Memento} w/o CBR) yields robust gains on each benchmark (HLE: $+11.0$ / $+1.6$, SimpleQA: $+32.5$ / $+4.9$, DeepResearcher: $+29.1$ / $+11.5$), indicating that explicit decomposition and tool orchestration systematically improve execution. Finally, case-based reasoning provides consistent, additive improvements (HLE: $+4.5$ / $+7.0$, SimpleQA: $+3.7$ / $+5.3$, DeepResearcher: $+6.7$ / $+8.2$). For HLE, however, without sufficient domain knowledge encoded in the backbone model, neither tool usage nor planning alone can reliably produce correct answers on long-tail, expert-level tasks. For DeepResearcher, we also identify data contamination~\citep{shumailov2024ai} across the seven evaluated benchmarks, evidenced by a noticeable drop in both F$1$ and PM when moving from the offline executor to the online executor without planning (DeepResearcher: $-18.0$ F$1$ / $-2.1$ PM). This aligns with broader findings in the field~\citep{sun2022recitation, yu2022generate, zhou2025trustrag}: simply using external knowledge can sometimes negatively affect the model, while the internal knowledge within the model plays an important role in QA tasks and can even outperform RAG.

\subsubsection{Continual Learning Ability Boosted by Parametric and Non-Parametric CBR} 

Figure ~\ref{fig:iteration} and Table~\ref{table:ec} present the continual learning curves across different memory designs for the \emph{Memento} framework, comparing the performance of three configurations: \emph{Memento} with non-parametric CBR or parametric CBR and \emph{Memento} without CBR. The results demonstrate that the full \emph{Memento} architecture consistently outperforms the ablated versions across all iterations, achieving higher accuracy at each step. Notably, removing CBR leads to a noticeable decline in performance, highlighting the effectiveness and complementary benefits of both parametric CBR and non-parametric CBR components in enhancing the continual learning capability of \emph{Memento}. More importantly, we observe a learning curve of the accuracy on the DeepResearcher dataset with increased iterations, suggesting that memory-based approaches can effectively enhance LLM agents without requiring parameter updates.

Although we attempted to locate any performance drops along the learning curve, in practice, such inflexion points are elusive. With only about $3$k training data, the Case Bank saturates quickly. Each additional iteration, therefore, contains progressively fewer previously unseen (and thus potentially failing) cases. In our simulated, open-ended, but ultimately finite environment, we observe rapid convergence with only marginal gains after a few iterations. Consequently, adding many more iterations yields diminishing returns and contributes little to our understanding of memory-based continual learning.

\subsubsection{Generalisation across Tasks}

To assess out-of-distribution (OOD) generalisation, we follow the evaluation protocol of~\citet{zheng2025deepresearcher}. Specifically, MusiQue~\citep{trivedi2022musique}, Bamboogle~\citep{press2022measuring}, and PopQA~\citep{mallen2022not} are selected as OOD datasets due to their distinct question styles and information distributions, while NQ~\citep{kwiatkowski2019natural}, TQ~\citep{joshi2017triviaqa}, HotpotQA~\citep{yang2018hotpotqa}, and 2Wiki~\citep{ho2020constructing} are used for training. We first collect and store trajectories from the training datasets in the case bank. During inference, \emph{Memento} retrieves the four most relevant cases from the case bank for each target query. As shown in Figure~\ref{fig:ood}, \emph{Memento} achieves substantial improvements on all OOD benchmarks, with absolute gains ranging from 4.7\% to 9.6\%. These results highlight the effectiveness of case-based reasoning in enhancing generalisation to unseen tasks.

 \begin{figure}[t]
  \centering
  \includegraphics[width=\linewidth]{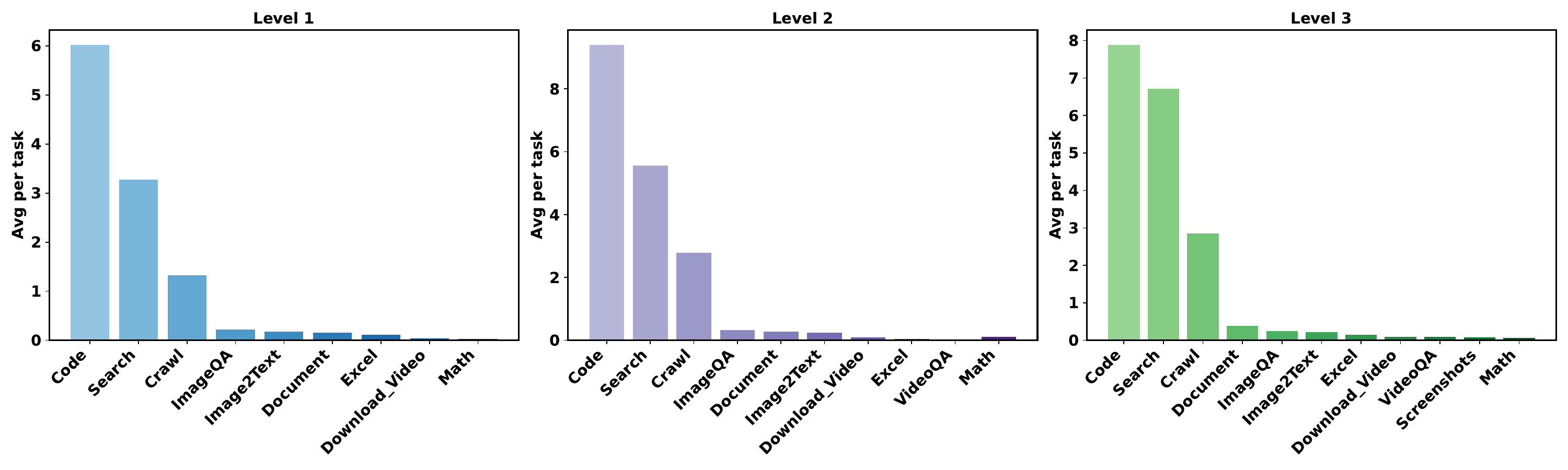}
  \caption{The average number of each task type per level, highlighting the dominance of code, search, and crawl tasks as difficulty level increases.}
  \label{fig:tools}
\end{figure}

\section{Discussion and Analysis}

Building on the results in \S~\ref{sec:experiments} that establish the effectiveness of \emph{Memento}, we further analyse its efficiency and operational behaviour. Specifically, we (i) analyse the average number of tool calls per task across three difficulty levels to assess how the MCP framework adapts as task complexity increases, (ii) characterise tool-call statistics, and (iii) evaluate the impact of using reasoning-oriented versus general-purpose models.

\subsection{The Number of Tokens per Task}

 As shown in Figure~\ref{fig:tools}, code, search, and crawl tasks dominate across all levels, with their usage increasing notably as difficulty rises. Importantly, while overall tool usage grows with task complexity, the most challenging problems increasingly rely on the model’s internal reasoning to interpret and aggregate evidence from prior tool outputs, rather than simply calling more tools via MCP. This highlights the importance of effective integration between planning, memory, and evidence aggregation for solving open-ended, long-horizon deep research tasks.

\subsection{Statistics of MCP Tools}

\begin{wrapfigure}{r}{0.33\linewidth}
  \centering
  \includegraphics[width=0.98\linewidth]{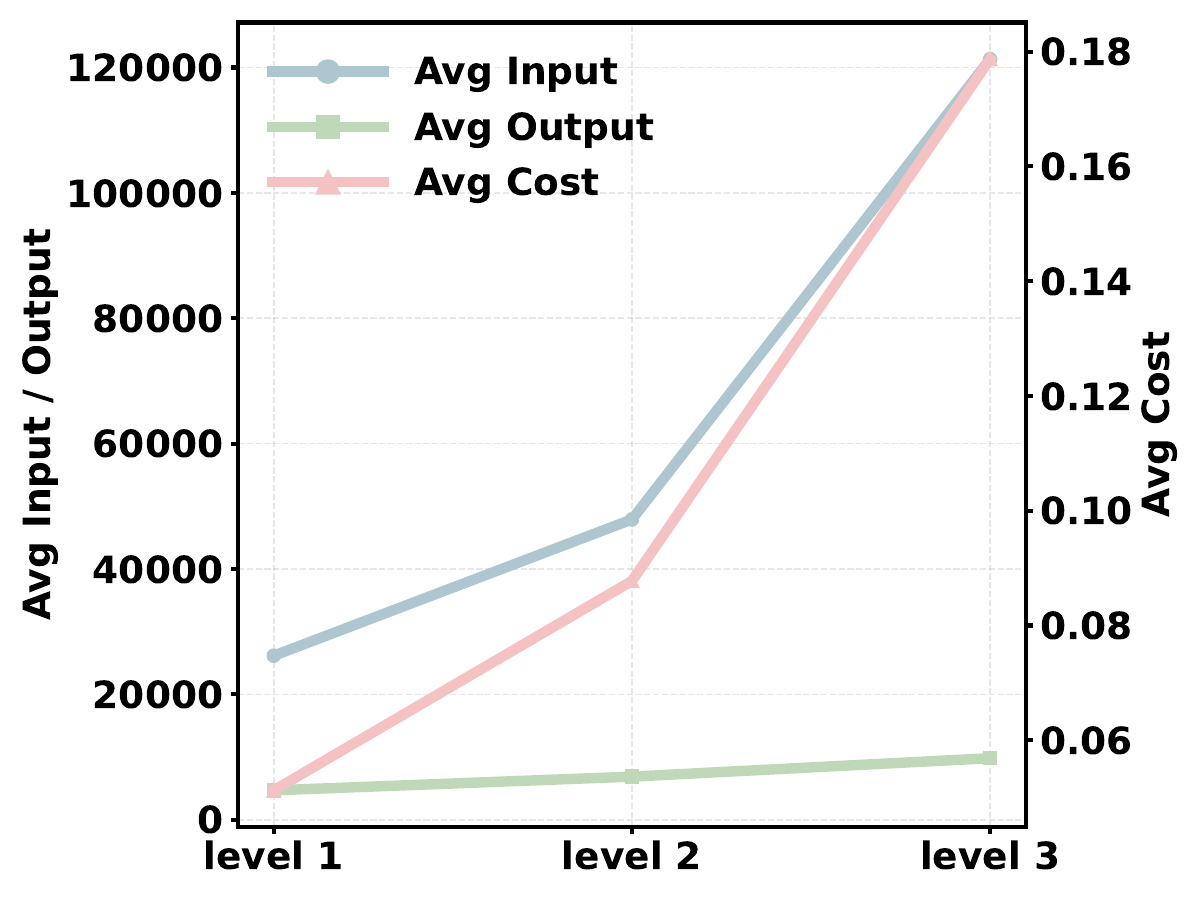}
  \caption{Token costs on the GAIA.}
  \label{fig:5}
  % \vspace{-5em}
\end{wrapfigure}

As shown in Figure~\ref{fig:5}, we randomly sample $10$ tasks from the GAIA validation, respectively, to calculate the tokens and costs per task. Average response tokens rise sharply with task difficulty: Level $1$ queries required $26 k$ input tokens and $4.7 k$ output tokens, Level $2$ grew to $48 k / 6.9 k$, and Level 3 peaked at $121 k / 9.8 k$. This highlights that the primary computational burden in complex scenarios stems from integrating and analysing multi-step tool outputs, rather than from generating long responses.

We observe that the output tokens remain stable across task levels, as the final answers typically require only short responses. This demonstrates that our system effectively controls generation length and avoids unnecessary verbosity during inference. However, due to the complexity and unpredictability of real-world environments, the input context grows significantly with task difficulty. As tasks become more complex, more detailed observations, plans, tool outputs, and intermediate reasoning steps must be incorporated into the input prompts, resulting in a substantial increase in the number of input tokens.

\subsection{The Impact of Fast and Slow Think Mode}

\begin{table}[t]
  \centering
  \label{tab:planner-executor}
  \begin{tabular}{@{}llrrrr@{}}
    \toprule
    Planner & Executor & Level 1 & Level 2 & Level 3 & Average \\
    \midrule
    gpt-4.1 & o3       & 77.36\% & 69.77\% & 61.54\% & 70.91\% \\
    o3 & o3       & 73.58\% & 63.95\% & 38.46\% & 63.03\% \\
    Qwen3-32B-Fast  & o4-mini  & 62.26\% & 56.98\% & 26.92\% & 53.94\% \\
    Qwen3-32B-Slow  & o4-mini  & 56.60\% & 36.05\% & 23.08\% & 40.61\% \\
    \bottomrule
  \end{tabular}
    \caption{The impact of fast and slow think mode on GAIA validation dataset (pass@1).}
    \label{table:fast-slow}
\end{table}

Table~\ref{table:fast-slow} compares the impact of fast- and slow-thinking planners on overall system performance (pass@1) across different task difficulties. The results show that pairing the fast, non-deliberative GPT-4.1 planner with the o3 executor yields the highest average accuracy (70.9\%), outperforming the more deliberative o3 planner (63.03\%) even when both use the same executor. Similarly, when using the o4-mini executor, GPT-4.1 achieves a substantial 16.4\% improvement over o3. The Qwen3-32B models further confirm this trend, with the fast planner consistently outperforming its slow counterpart.

Analysis of system traces reveals several key reasons for the fast planner’s superiority. The planner relying on the o3 model often either answers directly -- skipping plan generation altogether -- or produces overly verbose plans, which can mislead the executor with incomplete instructions. Additionally, in complex multi-step reasoning fields, the slow planner tends to compress solutions into a single, convoluted chain of thought, while the fast planner effectively decomposes problems into manageable sub-tasks.

Overall, these findings highlight that in modular LLM systems, concise and structured planning leads to more effective downstream execution. Overly deliberative planning not only introduces unnecessary context and redundancy but also induces role confusion, thereby undermining the very specialisation that the two-stage architecture is designed to exploit.

\section{Conclusion}

We introduce \emph{Memento}, a memory-based learning paradigm that enables LLM agents to adapt online search without updating model weights. \emph{Memento} formalises deep research agents as a memory-based Markov Decision Process (MDP) and implements it within a planner–executor framework, leveraging an episodic case bank to record and retrieve trajectories for continual policy improvement. Empirically, we achieve strong performance across GAIA, DeepResearcher, and SimpleQA. Ablation studies reveal that both parametric and non-parametric CBR are critical to the significant performance gains, and that a small, curated memory yields optimal results. These findings motivate future work on deep research tasks using memory-based MDP.

\bibliography{main}

\begin{thebibliography}{83}
\providecommand{\natexlab}[1]{#1}
\providecommand{\url}[1]{\texttt{#1}}
\expandafter\ifx\csname urlstyle\endcsname\relax
  \providecommand{\doi}[1]{doi: #1}\else
  \providecommand{\doi}{doi: \begingroup \urlstyle{rm}\Url}\fi

\bibitem[Aamodt and Plaza(1994)]{aamodt1994case}
Agnar Aamodt and Enric Plaza.
\newblock Case-based reasoning: Foundational issues, methodological variations, and system approaches.
\newblock \emph{AI communications}, 7\penalty0 (1):\penalty0 39--59, 1994.

\bibitem[Agarwal et~al.(2024)Agarwal, Singh, Zhang, Bohnet, Rosias, Chan, Zhang, Anand, Abbas, Nova, et~al.]{agarwal2024many}
Rishabh Agarwal, Avi Singh, Lei Zhang, Bernd Bohnet, Luis Rosias, Stephanie Chan, Biao Zhang, Ankesh Anand, Zaheer Abbas, Azade Nova, et~al.
\newblock Many-shot in-context learning.
\newblock \emph{Advances in Neural Information Processing Systems}, 37:\penalty0 76930--76966, 2024.

\bibitem[Alibaba(2025)]{aworld2025}
Alibaba.
\newblock Aworld: A unified agent playground for computer and phone use tasks, 2025.
\newblock URL \url{https://github.com/inclusionAI/AWorld}.

\bibitem[Anderson(2013)]{anderson2013architecture}
John~R Anderson.
\newblock \emph{The architecture of cognition}.
\newblock Psychology Press, 2013.

\bibitem[Anderson et~al.(1997)Anderson, Matessa, and Lebiere]{anderson1997act}
John~R Anderson, Michael Matessa, and Christian Lebiere.
\newblock Act-r: A theory of higher level cognition and its relation to visual attention.
\newblock \emph{Human--Computer Interaction}, 12\penalty0 (4):\penalty0 439--462, 1997.

\bibitem[Ashley(1992)]{ashley1992case}
Kevin~D Ashley.
\newblock Case-based reasoning and its implications for legal expert systems.
\newblock \emph{Artificial Intelligence and Law}, 1\penalty0 (2):\penalty0 113--208, 1992.

\bibitem[Baddeley(1983)]{baddeley1983working}
Alan~David Baddeley.
\newblock Working memory.
\newblock \emph{Philosophical Transactions of the Royal Society of London. B, Biological Sciences}, 302\penalty0 (1110):\penalty0 311--324, 1983.

\bibitem[{ByteDance}(2025)]{bytedance2025deerflow}
{ByteDance}.
\newblock Deerflow: Deep exploration and efficient research framework.
\newblock \url{https://deerflow.tech/z}, 2025.

\bibitem[Camel-AI(2025)]{owl2025}
Camel-AI.
\newblock Owl: Optimized workforce learning for general multi-agent assistance in real-world task automation, 2025.
\newblock URL \url{https://github.com/camel-ai/owl}.

\bibitem[Chhikara et~al.(2025)Chhikara, Khant, Aryan, Singh, and Yadav]{chhikara2025mem0}
Prateek Chhikara, Dev Khant, Saket Aryan, Taranjeet Singh, and Deshraj Yadav.
\newblock Mem0: Building production-ready ai agents with scalable long-term memory.
\newblock \emph{arXiv preprint arXiv:2504.19413}, 2025.

\bibitem[Choudhary et~al.(2021)Choudhary, Modani, and Maurya]{choudhary2021react}
Gautam Choudhary, Natwar Modani, and Nitish Maurya.
\newblock React: A review comment dataset for actionability (and more).
\newblock In \emph{Web Information Systems Engineering--WISE 2021: 22nd International Conference on Web Information Systems Engineering, WISE 2021, Melbourne, VIC, Australia, October 26--29, 2021, Proceedings, Part II 22}, pages 336--343. Springer, 2021.

\bibitem[Christianos et~al.(2023)Christianos, Papoudakis, Zimmer, Coste, Wu, Chen, Khandelwal, Doran, Feng, Liu, et~al.]{christianos2023pangu}
Filippos Christianos, Georgios Papoudakis, Matthieu Zimmer, Thomas Coste, Zhihao Wu, Jingxuan Chen, Khyati Khandelwal, James Doran, Xidong Feng, Jiacheng Liu, et~al.
\newblock Pangu-agent: A fine-tunable generalist agent with structured reasoning.
\newblock \emph{arXiv preprint arXiv:2312.14878}, 2023.

\bibitem[Cui et~al.(2021)Cui, Wang, Zhang, Chen, Luo, and Ooi]{cui2021alphaevolve}
Can Cui, Wei Wang, Meihui Zhang, Gang Chen, Zhaojing Luo, and Beng~Chin Ooi.
\newblock Alphaevolve: A learning framework to discover novel alphas in quantitative investment.
\newblock In \emph{Proceedings of the 2021 International conference on management of data}, pages 2208--2216, 2021.

\bibitem[Erdogan et~al.(2025)Erdogan, Lee, Kim, Moon, Furuta, Anumanchipalli, Keutzer, and Gholami]{erdogan2025plan}
Lutfi~Eren Erdogan, Nicholas Lee, Sehoon Kim, Suhong Moon, Hiroki Furuta, Gopala Anumanchipalli, Kurt Keutzer, and Amir Gholami.
\newblock Plan-and-act: Improving planning of agents for long-horizon tasks.
\newblock \emph{arXiv preprint arXiv:2503.09572}, 2025.

\bibitem[Feng et~al.(2025)Feng, Huang, Qu, Zhang, Qin, Zhong, Jiang, Chi, and Zhong]{feng2025retool}
Jiazhan Feng, Shijue Huang, Xingwei Qu, Ge~Zhang, Yujia Qin, Baoquan Zhong, Chengquan Jiang, Jinxin Chi, and Wanjun Zhong.
\newblock Retool: Reinforcement learning for strategic tool use in llms.
\newblock \emph{arXiv preprint arXiv:2504.11536}, 2025.

\bibitem[Fountas et~al.(2024)Fountas, Benfeghoul, Oomerjee, Christopoulou, Lampouras, Bou-Ammar, and Wang]{fountas2024human}
Zafeirios Fountas, Martin~A Benfeghoul, Adnan Oomerjee, Fenia Christopoulou, Gerasimos Lampouras, Haitham Bou-Ammar, and Jun Wang.
\newblock Human-like episodic memory for infinite context llms.
\newblock \emph{arXiv preprint arXiv:2407.09450}, 2024.

\bibitem[Francis and Ram(1993)]{francis1993utility}
Anthony~G Francis and Ashwin Ram.
\newblock The utility problem in case-based reasoning.
\newblock In \emph{Case-Based Reasoning: Papers from the 1993 Workshop}, pages 160--161, 1993.

\bibitem[Fu et~al.(2024)Fu, Kim, Kim, Sohn, Logeswaran, Bae, and Lee]{fu2024autoguide}
Yao Fu, Dong-Ki Kim, Jaekyeom Kim, Sungryull Sohn, Lajanugen Logeswaran, Kyunghoon Bae, and Honglak Lee.
\newblock Autoguide: Automated generation and selection of state-aware guidelines for large language model agents.
\newblock \emph{CoRR}, 2024.

\bibitem[Gao et~al.(2023)Gao, Xiong, Gao, Jia, Pan, Bi, Dai, Sun, Wang, and Wang]{gao2023retrieval}
Yunfan Gao, Yun Xiong, Xinyu Gao, Kangxiang Jia, Jinliu Pan, Yuxi Bi, Yixin Dai, Jiawei Sun, Haofen Wang, and Haofen Wang.
\newblock Retrieval-augmented generation for large language models: A survey.
\newblock \emph{arXiv preprint arXiv:2312.10997}, 2\penalty0 (1), 2023.

\bibitem[Glimcher(2011)]{glimcher2011understanding}
Paul~W Glimcher.
\newblock Understanding dopamine and reinforcement learning: the dopamine reward prediction error hypothesis.
\newblock \emph{Proceedings of the National Academy of Sciences}, 108\penalty0 (supplement\_3):\penalty0 15647--15654, 2011.

\bibitem[{Google}(2025)]{google2025deepresearch}
{Google}.
\newblock Gemini deep research — your personal research assistant.
\newblock \url{https://gemini.google/overview/deep-research/?hl=en-GB}, 2025.

\bibitem[Grosnit et~al.(2024)Grosnit, Maraval, Doran, Paolo, Thomas, Beevi, Gonzalez, Khandelwal, Iacobacci, Benechehab, et~al.]{grosnit2024large}
Antoine Grosnit, Alexandre Maraval, James Doran, Giuseppe Paolo, Albert Thomas, Refinath Shahul Hameed~Nabeezath Beevi, Jonas Gonzalez, Khyati Khandelwal, Ignacio Iacobacci, Abdelhakim Benechehab, et~al.
\newblock Large language models orchestrating structured reasoning achieve kaggle grandmaster level.
\newblock \emph{arXiv preprint arXiv:2411.03562}, 2024.

\bibitem[Guo et~al.(2024)Guo, Deng, Wen, Chen, Chang, and Wang]{guo2024ds}
Siyuan Guo, Cheng Deng, Ying Wen, Hechang Chen, Yi~Chang, and Jun Wang.
\newblock Ds-agent: Automated data science by empowering large language models with case-based reasoning.
\newblock In \emph{International Conference on Machine Learning (ICML)}, pages 16813--16848. PMLR, 2024.

\bibitem[Guo et~al.(2025)Guo, Liu, Chen, Xie, Zhang, Han, Chen, Chang, and Wang]{guo2025optimizing}
Siyuan Guo, Huiwu Liu, Xiaolong Chen, Yuming Xie, Liang Zhang, Tao Han, Hechang Chen, Yi~Chang, and Jun Wang.
\newblock Optimizing case-based reasoning system for functional test script generation with large language models.
\newblock In \emph{Proceedings of the 31st ACM SIGKDD Conference on Knowledge Discovery and Data Mining}, pages 4487--4498, 2025.

\bibitem[Haarnoja et~al.(2017)Haarnoja, Tang, Abbeel, and Levine]{haarnoja2017reinforcement}
Tuomas Haarnoja, Haoran Tang, Pieter Abbeel, and Sergey Levine.
\newblock Reinforcement learning with deep energy-based policies.
\newblock In \emph{International conference on machine learning}, pages 1352--1361. PMLR, 2017.

\bibitem[Haarnoja et~al.(2018)Haarnoja, Zhou, Hartikainen, Tucker, Ha, Tan, Kumar, Zhu, Gupta, Abbeel, et~al.]{haarnoja2018soft}
Tuomas Haarnoja, Aurick Zhou, Kristian Hartikainen, George Tucker, Sehoon Ha, Jie Tan, Vikash Kumar, Henry Zhu, Abhishek Gupta, Pieter Abbeel, et~al.
\newblock Soft actor-critic algorithms and applications.
\newblock \emph{arXiv preprint arXiv:1812.05905}, 2018.

\bibitem[Hatalis et~al.(2025)Hatalis, Christou, and Kondapalli]{hatalis2025review}
Kostas Hatalis, Despina Christou, and Vyshnavi Kondapalli.
\newblock Review of case-based reasoning for llm agents: theoretical foundations, architectural components, and cognitive integration.
\newblock \emph{arXiv preprint arXiv:2504.06943}, 2025.

\bibitem[Ho et~al.(2020)Ho, Nguyen, Sugawara, and Aizawa]{ho2020constructing}
Xanh Ho, Anh-Khoa~Duong Nguyen, Saku Sugawara, and Akiko Aizawa.
\newblock Constructing a multi-hop qa dataset for comprehensive evaluation of reasoning steps.
\newblock \emph{arXiv preprint arXiv:2011.01060}, 2020.

\bibitem[Huang et~al.(2025)Huang, Chen, Zhang, Li, Fang, Yang, Li, Shang, Xu, Hao, Kun, and Wang]{huang2025deep}
Yuxuan Huang, Yihang Chen, Haozheng Zhang, Kang Li, Meng Fang, Linyi Yang, Xiaoguang Li, Lifeng Shang, Songcen Xu, Jianye Hao, Shao Kun, and Jun Wang.
\newblock Deep research agents: A systematic examination and roadmap.
\newblock \emph{arXiv preprint arXiv:2506.18096}, 2025.

\bibitem[Jin et~al.(2025)Jin, Zeng, Yue, Yoon, Arik, Wang, Zamani, and Han]{jin2025search}
Bowen Jin, Hansi Zeng, Zhenrui Yue, Jinsung Yoon, Sercan Arik, Dong Wang, Hamed Zamani, and Jiawei Han.
\newblock Search-r1: Training llms to reason and leverage search engines with reinforcement learning.
\newblock \emph{arXiv preprint arXiv:2503.09516}, 2025.

\bibitem[Joshi et~al.(2017)Joshi, Choi, Weld, and Zettlemoyer]{joshi2017triviaqa}
Mandar Joshi, Eunsol Choi, Daniel~S Weld, and Luke Zettlemoyer.
\newblock Triviaqa: A large scale distantly supervised challenge dataset for reading comprehension.
\newblock \emph{arXiv preprint arXiv:1705.03551}, 2017.

\bibitem[Khosla et~al.(2023)Khosla, Zhu, and He]{khosla2023survey}
Savya Khosla, Zhen Zhu, and Yifei He.
\newblock Survey on memory-augmented neural networks: Cognitive insights to ai applications.
\newblock \emph{arXiv preprint arXiv:2312.06141}, 2023.

\bibitem[Kwiatkowski et~al.(2019)Kwiatkowski, Palomaki, Redfield, Collins, Parikh, Alberti, Epstein, Polosukhin, Devlin, Lee, et~al.]{kwiatkowski2019natural}
Tom Kwiatkowski, Jennimaria Palomaki, Olivia Redfield, Michael Collins, Ankur Parikh, Chris Alberti, Danielle Epstein, Illia Polosukhin, Jacob Devlin, Kenton Lee, et~al.
\newblock Natural questions: a benchmark for question answering research.
\newblock \emph{Transactions of the Association for Computational Linguistics}, 7:\penalty0 453--466, 2019.

\bibitem[Lewis et~al.(2020)Lewis, Perez, Piktus, Petroni, Karpukhin, Goyal, K{\"u}ttler, Lewis, Yih, Rockt{\"a}schel, et~al.]{lewis2020retrieval}
Patrick Lewis, Ethan Perez, Aleksandra Piktus, Fabio Petroni, Vladimir Karpukhin, Naman Goyal, Heinrich K{\"u}ttler, Mike Lewis, Wen-tau Yih, Tim Rockt{\"a}schel, et~al.
\newblock Retrieval-augmented generation for knowledge-intensive nlp tasks.
\newblock \emph{Advances in neural information processing systems}, 33:\penalty0 9459--9474, 2020.

\bibitem[Li et~al.(2025{\natexlab{a}})Li, Xue, Zhang, Yang, Zhang, Wang, Yu, Hui, Lin, and Liu]{li2025start}
Chengpeng Li, Mingfeng Xue, Zhenru Zhang, Jiaxi Yang, Beichen Zhang, Xiang Wang, Bowen Yu, Binyuan Hui, Junyang Lin, and Dayiheng Liu.
\newblock Start: Self-taught reasoner with tools.
\newblock \emph{arXiv preprint arXiv:2503.04625}, 2025{\natexlab{a}}.

\bibitem[Li et~al.(2024)Li, Ding, Fang, and Tao]{li2024revisiting}
Hongyu Li, Liang Ding, Meng Fang, and Dacheng Tao.
\newblock Revisiting catastrophic forgetting in large language model tuning.
\newblock \emph{arXiv preprint arXiv:2406.04836}, 2024.

\bibitem[Li et~al.(2025{\natexlab{b}})Li, Zhang, Yin, Zhang, Ou, Wu, Yin, Li, Tao, Wang, et~al.]{li2025websailor}
Kuan Li, Zhongwang Zhang, Huifeng Yin, Liwen Zhang, Litu Ou, Jialong Wu, Wenbiao Yin, Baixuan Li, Zhengwei Tao, Xinyu Wang, et~al.
\newblock Websailor: Navigating super-human reasoning for web agent.
\newblock \emph{arXiv preprint arXiv:2507.02592}, 2025{\natexlab{b}}.

\bibitem[Li et~al.(2023)Li, Zhao, Yu, Song, Li, Yu, Li, Huang, and Li]{li2023api}
Minghao Li, Yingxiu Zhao, Bowen Yu, Feifan Song, Hangyu Li, Haiyang Yu, Zhoujun Li, Fei Huang, and Yongbin Li.
\newblock Api-bank: A comprehensive benchmark for tool-augmented llms.
\newblock \emph{arXiv preprint arXiv:2304.08244}, 2023.

\bibitem[Li et~al.(2025{\natexlab{c}})Li, Dong, Jin, Zhang, Zhou, Zhu, Zhang, and Dou]{li2025search}
Xiaoxi Li, Guanting Dong, Jiajie Jin, Yuyao Zhang, Yujia Zhou, Yutao Zhu, Peitian Zhang, and Zhicheng Dou.
\newblock Search-o1: Agentic search-enhanced large reasoning models.
\newblock \emph{arXiv preprint arXiv:2501.05366}, 2025{\natexlab{c}}.

\bibitem[Liang et~al.(2025)Liang, Xiang, Yu, Zhang, Hong, Fan, and Tang]{openmanus2025}
Xinbin Liang, Jinyu Xiang, Zhaoyang Yu, Jiayi Zhang, Sirui Hong, Sheng Fan, and Xiao Tang.
\newblock Openmanus: An open-source framework for building general ai agents, 2025.
\newblock URL \url{https://doi.org/10.5281/zenodo.15186407}.

\bibitem[Liang et~al.(2024)Liang, He, Xia, Song, Wang, Tao, Sun, Yuan, Su, Li, et~al.]{liang2024self}
Xuechen Liang, Yangfan He, Yinghui Xia, Xinyuan Song, Jianhui Wang, Meiling Tao, Li~Sun, Xinhang Yuan, Jiayi Su, Keqin Li, et~al.
\newblock Self-evolving agents with reflective and memory-augmented abilities.
\newblock \emph{arXiv preprint arXiv:2409.00872}, 2024.

\bibitem[Liu et~al.(2025)Liu, Chen, Li, Qi, Pang, Du, Lee, and Lin]{liu2025understanding}
Zichen Liu, Changyu Chen, Wenjun Li, Penghui Qi, Tianyu Pang, Chao Du, Wee~Sun Lee, and Min Lin.
\newblock Understanding r1-zero-like training: A critical perspective.
\newblock \emph{arXiv preprint arXiv:2503.20783}, 2025.

\bibitem[Mallen et~al.(2022)Mallen, Asai, Zhong, Das, Hajishirzi, and Khashabi]{mallen2022not}
Alex Mallen, Akari Asai, Victor Zhong, Rajarshi Das, Hannaneh Hajishirzi, and Daniel Khashabi.
\newblock When not to trust language models: Investigating effectiveness and limitations of parametric and non-parametric memories.
\newblock \emph{arXiv preprint arXiv:2212.10511}, 7, 2022.

\bibitem[Mialon et~al.(2023)Mialon, Fourrier, Wolf, LeCun, and Scialom]{mialon2023gaia}
Gr{\'e}goire Mialon, Cl{\'e}mentine Fourrier, Thomas Wolf, Yann LeCun, and Thomas Scialom.
\newblock Gaia: a benchmark for general ai assistants.
\newblock In \emph{The Twelfth International Conference on Learning Representations}, 2023.

\bibitem[Nakano et~al.(2021)Nakano, Hilton, Balaji, Wu, Ouyang, Kim, Hesse, Jain, Kosaraju, Saunders, et~al.]{nakano2021webgpt}
Reiichiro Nakano, Jacob Hilton, Suchir Balaji, Jeff Wu, Long Ouyang, Christina Kim, Christopher Hesse, Shantanu Jain, Vineet Kosaraju, William Saunders, et~al.
\newblock Webgpt: Browser-assisted question-answering with human feedback.
\newblock \emph{arXiv preprint arXiv:2112.09332}, 2021.

\bibitem[{OpenAI}(2025)]{openai2025deepresearch}
{OpenAI}.
\newblock Deep research system card.
\newblock Technical report, OpenAI, 2025.
\newblock URL \url{https://cdn.openai.com/deep-research-system-card.pdf}.

\bibitem[Phan et~al.(2025)Phan, Gatti, Han, Li, Hu, Zhang, Zhang, Shaaban, Ling, Shi, et~al.]{phan2025humanity}
Long Phan, Alice Gatti, Ziwen Han, Nathaniel Li, Josephina Hu, Hugh Zhang, Chen Bo~Calvin Zhang, Mohamed Shaaban, John Ling, Sean Shi, et~al.
\newblock Humanity's last exam.
\newblock \emph{arXiv preprint arXiv:2501.14249}, 2025.

\bibitem[Press et~al.(2022)Press, Zhang, Min, Schmidt, Smith, and Lewis]{press2022measuring}
Ofir Press, Muru Zhang, Sewon Min, Ludwig Schmidt, Noah~A Smith, and Mike Lewis.
\newblock Measuring and narrowing the compositionality gap in language models.
\newblock \emph{arXiv preprint arXiv:2210.03350}, 2022.

\bibitem[Pritzel et~al.(2017)Pritzel, Uria, Srinivasan, Badia, Vinyals, Hassabis, Wierstra, and Blundell]{pritzel2017neural}
Alexander Pritzel, Benigno Uria, Sriram Srinivasan, Adria~Puigdomenech Badia, Oriol Vinyals, Demis Hassabis, Daan Wierstra, and Charles Blundell.
\newblock Neural episodic control.
\newblock In \emph{International conference on machine learning}, pages 2827--2836. PMLR, 2017.

\bibitem[Qian et~al.(2025)Qian, Acikgoz, He, Wang, Chen, Hakkani-T{\"u}r, Tur, and Ji]{qian2025toolrl}
Cheng Qian, Emre~Can Acikgoz, Qi~He, Hongru Wang, Xiusi Chen, Dilek Hakkani-T{\"u}r, Gokhan Tur, and Heng Ji.
\newblock Toolrl: Reward is all tool learning needs.
\newblock \emph{arXiv preprint arXiv:2504.13958}, 2025.

\bibitem[Qiu et~al.(2025)Qiu, Qi, Zhang, Juan, Guo, Lu, Wang, Yao, Ren, Jiang, et~al.]{qiu2025alita}
Jiahao Qiu, Xuan Qi, Tongcheng Zhang, Xinzhe Juan, Jiacheng Guo, Yifu Lu, Yimin Wang, Zixin Yao, Qihan Ren, Xun Jiang, et~al.
\newblock Alita: Generalist agent enabling scalable agentic reasoning with minimal predefinition and maximal self-evolution.
\newblock \emph{arXiv preprint arXiv:2505.20286}, 2025.

\bibitem[Ross(1989)]{ross1989some}
Brian~H Ross.
\newblock Some psychological results on case-based reasoning.
\newblock In \emph{Proceedings: Case-based reasoning workshop}, pages 144--147. Morgan Kaufmann, 1989.

\bibitem[Salama et~al.(2025)Salama, Cai, Yuan, Currey, Sunkara, Zhang, and Benajiba]{salama2025meminsight}
Rana Salama, Jason Cai, Michelle Yuan, Anna Currey, Monica Sunkara, Yi~Zhang, and Yassine Benajiba.
\newblock Meminsight: Autonomous memory augmentation for llm agents.
\newblock \emph{arXiv preprint arXiv:2503.21760}, 2025.

\bibitem[Schick et~al.(2023)Schick, Dwivedi-Yu, Dess{\`\i}, Raileanu, Lomeli, Hambro, Zettlemoyer, Cancedda, and Scialom]{schick2023toolformer}
Timo Schick, Jane Dwivedi-Yu, Roberto Dess{\`\i}, Roberta Raileanu, Maria Lomeli, Eric Hambro, Luke Zettlemoyer, Nicola Cancedda, and Thomas Scialom.
\newblock Toolformer: Language models can teach themselves to use tools.
\newblock \emph{Advances in Neural Information Processing Systems}, 36:\penalty0 68539--68551, 2023.

\bibitem[Shi et~al.(2025)Shi, Tan, Kuang, Li, Ren, Zhang, Chen, Wang, Shang, Yu, et~al.]{shi2025pangu}
Wenxuan Shi, Haochen Tan, Chuqiao Kuang, Xiaoguang Li, Xiaozhe Ren, Chen Zhang, Hanting Chen, Yasheng Wang, Lifeng Shang, Fisher Yu, et~al.
\newblock Pangu deepdiver: Adaptive search intensity scaling via open-web reinforcement learning.
\newblock \emph{arXiv preprint arXiv:2505.24332}, 2025.

\bibitem[Shinn et~al.(2023)Shinn, Cassano, Gopinath, Narasimhan, and Yao]{shinn2023reflexion}
Noah Shinn, Federico Cassano, Ashwin Gopinath, Karthik Narasimhan, and Shunyu Yao.
\newblock Reflexion: Language agents with verbal reinforcement learning.
\newblock \emph{Advances in Neural Information Processing Systems}, 36:\penalty0 8634--8652, 2023.

\bibitem[Shumailov et~al.(2024)Shumailov, Shumaylov, Zhao, Papernot, Anderson, and Gal]{shumailov2024ai}
Ilia Shumailov, Zakhar Shumaylov, Yiren Zhao, Nicolas Papernot, Ross Anderson, and Yarin Gal.
\newblock Ai models collapse when trained on recursively generated data.
\newblock \emph{Nature}, 631\penalty0 (8022):\penalty0 755--759, 2024.

\bibitem[Smyth and McClave(2001)]{smyth2001similarity}
Barry Smyth and Paul McClave.
\newblock Similarity vs. diversity.
\newblock In \emph{International conference on case-based reasoning}, pages 347--361. Springer, 2001.

\bibitem[Song et~al.(2025)Song, Jiang, Min, Chen, Chen, Zhao, Fang, and Wen]{song2025r1}
Huatong Song, Jinhao Jiang, Yingqian Min, Jie Chen, Zhipeng Chen, Wayne~Xin Zhao, Lei Fang, and Ji-Rong Wen.
\newblock R1-searcher: Incentivizing the search capability in llms via reinforcement learning.
\newblock \emph{arXiv preprint arXiv:2503.05592}, 2025.

\bibitem[Squire et~al.(2015)Squire, Genzel, Wixted, and Morris]{squire2015memory}
Larry~R Squire, Lisa Genzel, John~T Wixted, and Richard~G Morris.
\newblock Memory consolidation.
\newblock \emph{Cold Spring Harbor perspectives in biology}, 7\penalty0 (8):\penalty0 a021766, 2015.

\bibitem[Sumers et~al.(2023)Sumers, Yao, Narasimhan, and Griffiths]{sumers2023cognitive}
Theodore Sumers, Shunyu Yao, Karthik Narasimhan, and Thomas Griffiths.
\newblock Cognitive architectures for language agents.
\newblock \emph{Transactions on Machine Learning Research}, 2023.

\bibitem[Sun et~al.(2022)Sun, Wang, Tay, Yang, and Zhou]{sun2022recitation}
Zhiqing Sun, Xuezhi Wang, Yi~Tay, Yiming Yang, and Denny Zhou.
\newblock Recitation-augmented language models.
\newblock \emph{arXiv preprint arXiv:2210.01296}, 2022.

\bibitem[Tang et~al.(2025)Tang, Qin, Peng, Zhou, Shao, Du, Wei, Xia, Wu, Zhu, et~al.]{tang2025agent}
Xiangru Tang, Tianrui Qin, Tianhao Peng, Ziyang Zhou, Daniel Shao, Tingting Du, Xinming Wei, Peng Xia, Fang Wu, He~Zhu, et~al.
\newblock Agent kb: Leveraging cross-domain experience for agentic problem solving.
\newblock \emph{arXiv preprint arXiv:2507.06229}, 2025.

\bibitem[Trivedi et~al.(2022)Trivedi, Balasubramanian, Khot, and Sabharwal]{trivedi2022musique}
Harsh Trivedi, Niranjan Balasubramanian, Tushar Khot, and Ashish Sabharwal.
\newblock Musique: Multihop questions via single-hop question composition.
\newblock \emph{Transactions of the Association for Computational Linguistics}, 10:\penalty0 539--554, 2022.

\bibitem[Wang et~al.(2025)Wang, Qian, Zhong, Chen, Qiu, Huang, Jin, Wang, Wong, and Ji]{wang2025otc}
Hongru Wang, Cheng Qian, Wanjun Zhong, Xiusi Chen, Jiahao Qiu, Shijue Huang, Bowen Jin, Mengdi Wang, Kam-Fai Wong, and Heng Ji.
\newblock Otc: Optimal tool calls via reinforcement learning.
\newblock \emph{arXiv preprint arXiv:2504.14870}, 2025.

\bibitem[Wang et~al.(2024)Wang, Mao, Fried, and Neubig]{wang2024agent}
Zora~Zhiruo Wang, Jiayuan Mao, Daniel Fried, and Graham Neubig.
\newblock Agent workflow memory.
\newblock \emph{arXiv preprint arXiv:2409.07429}, 2024.

\bibitem[Wei et~al.(2022)Wei, Wang, Schuurmans, Bosma, Xia, Chi, Le, Zhou, et~al.]{wei2022chain}
Jason Wei, Xuezhi Wang, Dale Schuurmans, Maarten Bosma, Fei Xia, Ed~Chi, Quoc~V Le, Denny Zhou, et~al.
\newblock Chain-of-thought prompting elicits reasoning in large language models.
\newblock \emph{Advances in neural information processing systems}, 35:\penalty0 24824--24837, 2022.

\bibitem[Wei et~al.(2024)Wei, Karina, Chung, Jiao, Papay, Glaese, Schulman, and Fedus]{wei2024measuring}
Jason Wei, Nguyen Karina, Hyung~Won Chung, Yunxin~Joy Jiao, Spencer Papay, Amelia Glaese, John Schulman, and William Fedus.
\newblock Measuring short-form factuality in large language models.
\newblock \emph{arXiv preprint arXiv:2411.04368}, 2024.

\bibitem[Weng et~al.(2025)Weng, Zhu, Bao, Zhang, Wang, Zhang, and Yang]{weng2025cycleresearcher}
Yixuan Weng, Minjun Zhu, Guangsheng Bao, Hongbo Zhang, Jindong Wang, Yue Zhang, and Linyi Yang.
\newblock Cycleresearcher: Improving automated research via automated review.
\newblock In \emph{The Thirteenth International Conference on Learning Representations}, 2025.
\newblock URL \url{https://openreview.net/forum?id=bjcsVLoHYs}.

\bibitem[Wiratunga et~al.(2024)Wiratunga, Abeyratne, Jayawardena, Martin, Massie, Nkisi-Orji, Weerasinghe, Liret, and Fleisch]{wiratunga2024cbr}
Nirmalie Wiratunga, Ramitha Abeyratne, Lasal Jayawardena, Kyle Martin, Stewart Massie, Ikechukwu Nkisi-Orji, Ruvan Weerasinghe, Anne Liret, and Bruno Fleisch.
\newblock Cbr-rag: case-based reasoning for retrieval augmented generation in llms for legal question answering.
\newblock In \emph{International Conference on Case-Based Reasoning}, pages 445--460. Springer, 2024.

\bibitem[Wu et~al.(2023)Wu, Bansal, Zhang, Wu, Li, Zhu, Jiang, Zhang, Zhang, Liu, et~al.]{wu2023autogen}
Qingyun Wu, Gagan Bansal, Jieyu Zhang, Yiran Wu, Beibin Li, Erkang Zhu, Li~Jiang, Xiaoyun Zhang, Shaokun Zhang, Jiale Liu, et~al.
\newblock Autogen: Enabling next-gen llm applications via multi-agent conversation.
\newblock \emph{arXiv preprint arXiv:2308.08155}, 2023.

\bibitem[Xu et~al.(2025)Xu, Mei, Gao, Tan, Liang, and Zhang]{xu2025mem}
Wujiang Xu, Kai Mei, Hang Gao, Juntao Tan, Zujie Liang, and Yongfeng Zhang.
\newblock A-mem: Agentic memory for llm agents.
\newblock \emph{arXiv preprint arXiv:2502.12110}, 2025.

\bibitem[Yan et~al.(2025)Yan, Song, Feng, Yang, Zhang, Ammar, and Wang]{yan2024efficient}
Xue Yan, Yan Song, Xidong Feng, Mengyue Yang, Haifeng Zhang, Haitham~Bou Ammar, and Jun Wang.
\newblock Efficient reinforcement learning with large language model priors.
\newblock \emph{The Thirteenth International Conference on Learning Representations(ICLR)}, 2025.

\bibitem[Yang et~al.(2025)Yang, Ma, Huang, Chai, Gong, Geng, Zhou, Wen, Fang, Chen, et~al.]{yang2025agentic}
Yingxuan Yang, Mulei Ma, Yuxuan Huang, Huacan Chai, Chenyu Gong, Haoran Geng, Yuanjian Zhou, Ying Wen, Meng Fang, Muhao Chen, et~al.
\newblock Agentic web: Weaving the next web with ai agents.
\newblock \emph{arXiv preprint arXiv:2507.21206}, 2025.

\bibitem[Yang et~al.(2018)Yang, Qi, Zhang, Bengio, Cohen, Salakhutdinov, and Manning]{yang2018hotpotqa}
Zhilin Yang, Peng Qi, Saizheng Zhang, Yoshua Bengio, William~W Cohen, Ruslan Salakhutdinov, and Christopher~D Manning.
\newblock Hotpotqa: A dataset for diverse, explainable multi-hop question answering.
\newblock \emph{arXiv preprint arXiv:1809.09600}, 2018.

\bibitem[Yao et~al.(2023)Yao, Zhao, Yu, Du, Shafran, Narasimhan, and Cao]{yao2023}
Shunyu Yao, Jeffrey Zhao, Dian Yu, Nan Du, Izhak Shafran, Karthik Narasimhan, and Yuan Cao.
\newblock React: Synergizing reasoning and acting in language models.
\newblock In \emph{International Conference on Learning Representations (ICLR)}, 2023.

\bibitem[Yu et~al.(2022)Yu, Iter, Wang, Xu, Ju, Sanyal, Zhu, Zeng, and Jiang]{yu2022generate}
Wenhao Yu, Dan Iter, Shuohang Wang, Yichong Xu, Mingxuan Ju, Soumya Sanyal, Chenguang Zhu, Michael Zeng, and Meng Jiang.
\newblock Generate rather than retrieve: Large language models are strong context generators.
\newblock \emph{arXiv preprint arXiv:2209.10063}, 2022.

\bibitem[Zhao et~al.(2024)Zhao, Huang, Xu, Lin, Liu, and Huang]{zhao2024expel}
Andrew Zhao, Daniel Huang, Quentin Xu, Matthieu Lin, Yong-Jin Liu, and Gao Huang.
\newblock Expel: Llm agents are experiential learners.
\newblock In \emph{Proceedings of the AAAI Conference on Artificial Intelligence}, volume~38, pages 19632--19642, 2024.

\bibitem[Zheng et~al.(2025)Zheng, Fu, Hu, Cai, Ye, Lu, and Liu]{zheng2025deepresearcher}
Yuxiang Zheng, Dayuan Fu, Xiangkun Hu, Xiaojie Cai, Lyumanshan Ye, Pengrui Lu, and Pengfei Liu.
\newblock Deepresearcher: Scaling deep research via reinforcement learning in real-world environments.
\newblock \emph{arXiv preprint arXiv:2504.03160}, 2025.

\bibitem[Zhong et~al.(2024)Zhong, Guo, Gao, Ye, and Wang]{zhong2024memorybank}
Wanjun Zhong, Lianghong Guo, Qiqi Gao, He~Ye, and Yanlin Wang.
\newblock Memorybank: Enhancing large language models with long-term memory.
\newblock In \emph{Proceedings of the AAAI Conference on Artificial Intelligence}, volume~38, pages 19724--19731, 2024.

\bibitem[Zhou et~al.(2025)Zhou, Lee, Zhan, Chen, and Li]{zhou2025trustrag}
Huichi Zhou, Kin-Hei Lee, Zhonghao Zhan, Yue Chen, and Zhenhao Li.
\newblock Trustrag: Enhancing robustness and trustworthiness in rag.
\newblock \emph{arXiv preprint arXiv:2501.00879}, 2025.

\bibitem[Zhu et~al.(2025{\natexlab{a}})Zhu, Weng, Yang, and Zhang]{zhu2025deepreview}
Minjun Zhu, Yixuan Weng, Linyi Yang, and Yue Zhang.
\newblock Deepreview: Improving llm-based paper review with human-like deep thinking process.
\newblock \emph{arXiv preprint arXiv:2503.08569}, 2025{\natexlab{a}}.

\bibitem[Zhu et~al.(2025{\natexlab{b}})Zhu, Xie, Weng, Wu, Lin, Yang, and Zhang]{zhu2025ai}
Minjun Zhu, Qiujie Xie, Yixuan Weng, Jian Wu, Zhen Lin, Linyi Yang, and Yue Zhang.
\newblock Ai scientists fail without strong implementation capability.
\newblock \emph{arXiv preprint arXiv:2506.01372}, 2025{\natexlab{b}}.

\end{thebibliography}

\newpage
\clearpage
\appendix
\section{Derivation of the Optimal Policy in Soft-Q Learning}
\label{app:der}
The soft value function over the state $s$ is defined as:
\iffalse
\begin{equation}
V^\pi(s|M)=\alpha \log \sum_{c' \in M} \exp(\frac{1}{\alpha}Q^\pi(s,c'|M)).
\end{equation}
\fi
\begin{equation}
V^{\pi}(s,M)=\sum_{c\in M}\mu(c|s,M)[Q^{\pi}(s,M,c)-\alpha\log\mu(c|s,M)]
\end{equation}
The Q function over the state, case pair is defined as:
\begin{equation}
Q^\pi(s,M,c)=\mathbb{E}_{a\sim p_{\text{LLM}}(\cdot|s,c), s'\sim\mathcal{P}(\cdot|s,a)} \left[ r(s,a) + \gamma V^\pi(s',M^\prime) \right],
\end{equation}

Define the visitation frequency over the state, case bank for the policy $\pi$ as: $d^{\pi}(s,M)=\sum_{t=0}^\infty\gamma^{t-1} \mathbb{P}(s_t=s,M_t=M)$.
%where $d^{\pi_\theta}$ denotes the state and case bank distribution over the policy $\pi_\theta$.
Then, our goal is to derive the optimal retrieval policy by expected value function:
\begin{equation}
\begin{aligned}
J_{\text{MaxEnt}}(\pi)&=\mathbb{E}_{(s,M)\sim d^{\pi}}\left[V^{\pi}(s,M)\right],\\
&= \mathbb{E}_{(s,M)\sim d^{\pi}}\left[\sum_{c\in M} \mu(c|s,M)\left[ Q^{\pi}(s,M,c)- \alpha \log\mu(c|s,M) \right]\right]
\end{aligned}
\end{equation}

For simplicity, let $\mu_c=\mu(c|s,M)$ and $Q_c=Q^{\pi}(s,M,c)$, and introduce the Lagrange multiplier $\lambda$ to constrain that $\sum_{c} \mu_c=1$. Then, for any state, case bank pair, we have the optimisation objective:
\begin{equation}    
\mathcal{J}(\{\mu_c\},\lambda)=\sum_{c} \mu_c Q_c - \alpha \sum_c \mu_c \log\mu_c - \lambda (\sum_c \mu_c - 1),
\end{equation}
whose derivative concerning $\mu_c$ is:
\begin{equation}    
\frac{\partial\mathcal{J}(\{\mu_c\},\lambda)}{\partial \mu_c}= Q_c - \alpha (1 + \log \mu_c) - \lambda,
\end{equation}
Let $\frac{\partial\mathcal{J}(\{\mu_c\},\lambda)}{\partial \mu_c}=0$, then we have:
\begin{equation}
\begin{aligned}
    \mu_c &= \exp(\frac{Q_c}{\alpha} - (\frac{\lambda}{\alpha}+1)) \\
    &=K\exp(\frac{Q_c}{\alpha}),
\end{aligned}
\end{equation}
where $K=\exp(-\frac{\lambda}{\alpha}-1)$. Thus, by performing the normalisation and shaping the optimal Q, we have:
\begin{equation}
\label{eq:retrieval-Q}
    \mu^*_c= \frac{\exp(Q^*_c/\alpha)}{\sum_{c'}\exp(Q^*_{c'}/\alpha)}.
\end{equation}
Note that when $\alpha \rightarrow 0$, the soft Q learning deteriorates to standard Q-learning. The softmax form of the policy is also used in previous LLM-based agents with LLM prior (\cite{yan2024efficient}).

\clearpage
\section{Analysis of Memory Mechanisms}
\label{app:mem_jus}
\begin{table}[H]
\centering
\small
\begin{tabular}{lcccccc}
\toprule
            Method & Kernel & Neural Q &Q-Function& Read & Write & Gradient \\ \midrule
Tabular Q-learning & w/o    & w/o   &  Q-Table & Exact Match   &   Eq.~\eqref{eq:TD}    & -    \\
Deep Q-learning              & w/o    & w/   &  Neural Network &   Eq.~\eqref{eq:softmax-Q}   &     Eq.~\eqref{eq:para_Q}  &  Eq.~\eqref{eq:grad_para_Q}    \\
Neural Episodic Control & w/     & w/o      &   Eq.~\eqref{eq:EC}   &  Eq.~\eqref{eq:softmax-Q} &  Eq.~\eqref{eq:final-td}  &   Eq.~\eqref{eq:final_grad}   \\
Non-Parametric Memory in Sec.~\ref{sec:implementation}              & w/o    & w/o   &  w/o &   Eq.~\eqref{eq:read_np}   &     Eq.~\eqref{eq:write}  &  -  \\
Parametric Memory in Sec.~\ref{sec:implementation}              & w/o    & w/   &  Neural Network &   Eq.~\eqref{eq:read_ec}   &     Eq.~\eqref{eq:dr_loss_ce}  &  Eq.~\eqref{eq:dr_grad_ce}    \\
\bottomrule
\end{tabular}
\caption{Detail comparison of memory mechanisms.}
\label{table:mem}
\end{table}

Here, we consider several representative memory mechanisms, emphasising their Read and Write operations as summarised in Table~\ref{table:mem}. Specifically, we discuss tabular and parametric Q-value representations, as well as EC-based methods.\\\\
In the tabular setting, the memory maintains an explicit table $Q: \mathcal{S}\times \mathcal{A} \rightarrow \mathbb{R}$, where Read is a direct lookup of table $Q(s,M,a)$ and Write corresponds to updating the entry for the state–action pair after observing a transition, following the standard TD learning in Eq.~\eqref{eq:TD}. To extend beyond discrete spaces, deep Q-learning learns the Q function by a neural network $Q(s,M,a;\theta)$, with Read operation sampling cases from the retrieval policy $\mu$ following Eq.~\eqref{eq:softmax-Q} and Write operation updates the parameters $\theta$ via minimising the TD error:
\begin{equation}
\label{eq:para_Q}
    \mathcal{L}(\theta)=\mathbb{E}_{\left(s, c, r, s^{\prime}, M, M^{\prime}\right)}\left[\left(Q(s, M, c ; \theta)-\left[r+\gamma \alpha \log \sum_{c^{\prime} \in M^{\prime}} \exp \left(Q\left(s^{\prime}, M^{\prime}, c^{\prime} ; \bar{\theta}\right)\right)\right]\right)^{2}\right],
\end{equation}
where $\bar{\theta}$ is the target Q network. The gradient of the TD learning loss with respect to $\theta$ is given by:
\begin{equation}
\label{eq:grad_para_Q}
\triangledown_{\theta} \mathcal{L}(\theta)=2 \mathbb{E}_{\left(s, c, r, s^{\prime}, M, M^{\prime}\right)}\left[(Q(s, M, c ; \theta)-y) \triangledown_{\theta} Q(s, M, c ; \theta)\right],
\end{equation}
where $y=r+\gamma\alpha\log\sum_{c'\in M'}\exp\!\big(Q(s^\prime,M^\prime,c^\prime;\bar{\theta})\big)$. This parametric formulation enables generalisation across states and action spaces through the shared parameters $\theta$, in contrast to tabular methods, which only memorise individual entries. However, this benefit comes at the cost of optimisation instability and large data demand, since approximation errors may propagate globally through the parameter space. This limitation motivates the EC-based methods in Section \ref{sec:methodology}, where value estimation is regularised through a learnable kernel (see Eq.~\eqref{eq:EC}). Within this memory design, the Read operation samples cases from the retrieval policy distribution defined in Eq.~\eqref{eq:softmax-Q} while the Write operation additionally store  $(s,c,Q)$ into an episodic memory and updates the kernel parameters $\theta$ by Eq.~\eqref{eq:final-td} with gradient in Eq.~\eqref{eq:final_grad} optimising the weighting function. This approach only parameterises the kernel to regularise the historical Q-values of matched states, thereby ensuring generalisation across the state space while retaining data-efficient adaptation and improved stability compared with deep Q-learning.\\\\
In section \ref{sec:implementation}, the CBR agent is implemented as a planner, applying both non-parametric and parametric memory mechanisms. For the non-parametric variant, the Write operation appends each observed case $(s_t,a_t,r_t)$ into the case bank as in Eq.\eqref{eq:write}, while the Read operation retrieves the most relevant experiences by performing cosine-similarity matching between the current query embedding and stored states, followed by a $\mathrm{TopK}$ selection as in Eq.~\eqref{eq:read_np}. This similarity-based retrieval, without further parameterisation, is a common design in CBR and provides an effective means of reusing past experiences. Alongside the non-parametric approach, the single-step nature of the deep research setting permits fitting a parametric Q-function directly, as the reduced state space substantially lowers data requirements.
% with the M-MDP simplified into a single-step setting. This simplification reduces the effective state space and thereby lowers data requirements, making it feasible to fit a parametric Q‑function directly with the available samples. 
In the single‑step case, the temporal‑difference bootstrap vanishes, so the learning objective reduces to Eq.\eqref{eq:dr_loss_mse}. Furthermore, since the reward signal in the deep research scenario is binary, we replace the MSE objective with a CE loss. This choice avoids the vanishing-gradient problem near the boundaries $0$ and $1$, while providing more numerically stable training signals. Consequently, the final updating objective is reformulated as a binary classification loss in Eq.\eqref{eq:dr_loss_ce}, and the resulting gradient is as follows:

\begin{equation}
\label{eq:dr_grad_ce}
    \triangledown_\theta\mathcal{L}(\theta) = \mathbb{E}_{(s,c,r)}\left[\frac{Q(s,c;\theta)-r}{Q(s,c;\theta)(1-Q(s,c;\theta))}\triangledown_\theta Q(s,c;\theta)
    \right].
\end{equation}
To further stabilise case selection, we also apply a $\mathrm{TopK}$ operator in the parametric Read operator Eq.~\eqref{eq:read_ec} rather than sampling from the retrieval policy $\mu$.

\end{document}